\documentclass{article}
\PassOptionsToPackage{numbers}{natbib}

\usepackage[preprint]{neurips_2026}

\usepackage[utf8]{inputenc} %
\usepackage[T1]{fontenc}    %
\usepackage{hyperref}       %
\usepackage{url}            %
\usepackage{booktabs}       %
\usepackage{amsfonts}       %
\usepackage{nicefrac}       %
\usepackage{microtype}      %
\usepackage{xcolor}         %
\usepackage{amsmath}
\usepackage{makecell}
\usepackage{graphicx}
\usepackage{subcaption}
\usepackage{mathtools}
\usepackage[normalem]{ulem}

\input{preamble}

\title{\texorpdfstring{\ourMethod}{RayTun3R}: \\Online Camera Adaptation in 3D Foundation Models}

\author{%
  Daniil Sinitsyn$^{1,2}$ \qquad
  Nikita Araslanov$^{1,2,3}$ \qquad
  Daniel Cremers$^{1,2}$\\[2mm]
  $^1$TU Munich\qquad
  $^2$Munich Center for Machine Learning\qquad
  $^3$University of Oxford
}

\begin{document}

\maketitle

\begin{abstract}
  Recent 3D foundation models, such as DUSt3R, MASt3R, VGGT, \pithree, and Depth Anything~3, provide strong feed-forward depth and pose estimates on pinhole imagery, but degrade sharply under fisheye camera geometry.
We show that this failure is partly caused by a pinhole camera bias in the positional encodings of pretrained 3D foundation models, and propose \ourMethod, a lightweight camera adaptation approach.
It keeps the pretrained network fixed and adapts only lightweight components tied to token position and camera geometry.
\ourMethod learns parameter-efficient residual corrections to absolute and rotary positional encodings, together with parameter-free tokenization and corrections to prediction-grid coordinates that remove residual pinhole assumptions. The resulting adapter contains only 10,752 trainable parameters and can be learned from a short temporal segment using geometric losses.
Once adapted, \ourMethod transfers effectively to the remaining frames of the sequence without incurring additional runtime costs.
Across diverse fisheye datasets with fields of view from $110^\circ$ to $200^\circ$, our adapter reduces rotation error by $2$--$12\times$ relative to the unadapted model, outperforms LoRA while using $\sim\!14\times$ fewer trainable parameters, improves pose over adaptation-free baselines while avoiding their multi-view inference cost, and remains competitive on depth accuracy.

\end{abstract}

\section{Introduction}
\label{sec:intro}

\begin{figure}[t]
    \centering
    \def\svgwidth{1.0\linewidth}
    \input{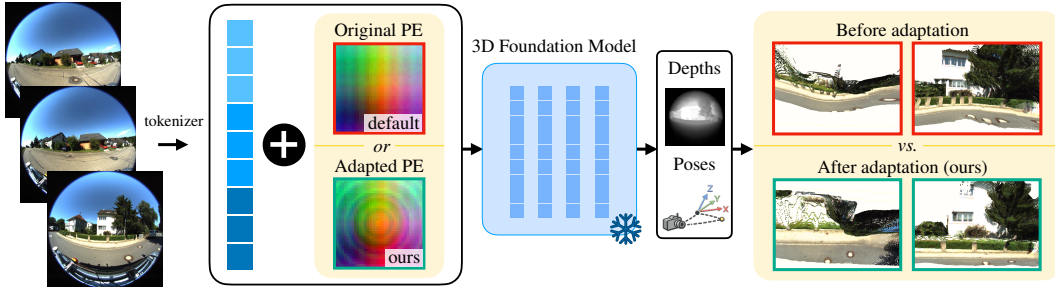}
    \caption{\textbf{\ourMethod overview.} The figure shows fisheye input, the original and adapted positional encodings, the frozen 3D foundation model, and the resulting geometry before and after adaptation. \ourMethod keeps the backbone frozen and adapts lightweight components tied to token location and camera geometry to recover more coherent depth, pose, and 3D structure.}

    \label{fig:da3_fisheye_failure}
\end{figure}

Large 3D foundation models, such as DUSt3R~\citep{wang2024dust3r}, MASt3R~\citep{duisterhof2025mastrsfm}, VGGT~\citep{wang2025vggt}, \pithree~\citep{wang2026pi3permutationequivariantvisualgeometry}, and Depth~Anything~3 (DA3)~\citep{lin2025depth}, can recover depth, camera poses, and coarse 3D structure from a small set of images in near-instant feed-forward fashion.
These models generalize well across everyday pinhole imagery; however, their robustness does not extend to changes in camera geometry. 
As \cref{fig:da3_fisheye_failure} shows, fisheye input degrades geometry across the image, including near the optical center.
A common workaround is a crop-based approach: Splitting the fisheye image into multiple virtual pinhole views~\citep{wang2018cubemapslam,xu2024sdge}, running inference with the foundation model on each view, and fusing the predictions. 
This allows the model to operate in a familiar input distribution, but also incurs additional computational costs and requires non-trivial prediction fusion without representing fisheye geometry directly.

We address this limitation with lightweight online adaptation of these pinhole-biased models.
Our key observation is that pinhole and fisheye images assign different ray geometry to the same image-grid locations, making positional embeddings a natural target for online adaptation.
By contrast, standard parameter-efficient fine-tuning (PEFT), such as LoRA~\citep{hu2022lora}, adapter layers~\citep{houlsby2019parameter}, SSF~\citep{lian2022scaling}, and calibration tokens~\citep{gangopadhyay2025extending}, adapts generic feature transformations without isolating the camera-induced  mismatch in the ray geometry.

Our adaptation method, \emph{\ourMethod}, keeps the pretrained 3D model frozen and updates only lightweight components that encode how token locations in the image relate to camera rays.
The learned parameters are small radial/angular lookup tables for absolute and rotary positional encodings; prediction-grid coordinates and patch-tokenization corrections are parameter-free.\footnote{We found that directly distorting the positional embedding table to be suboptimal, see \cref{sec:patch_undistort}.}
\ourMethod comes at no inference-time overhead and adds only 10,752 trainable parameters, learned without ground truth on a short temporal segment.
Across diverse fisheye datasets and three frozen backbones, it consistently improves pose and achieves competitive depth while using the full fisheye field of view.
Our code will be made publicly available.

\inparagraph{Contributions.}
\textit{(i)} We show that the local Jacobian of pretrained positional embeddings follows the geometry of the pinhole camera model, revealing a pinhole bias in the positional representation of 3D foundation models.
\textit{(ii)} Motivated by this analysis, we introduce a lightweight adapter for components tied to token locations in the image. It is runtime-efficient, adding only 10,752 trainable parameters on DA3-Small while keeping the pretrained model frozen.
\textit{(iii)} We empirically demonstrate that our adapter improves pose and depth accuracy of all foundation models tested (Depth Anything 3, VGGT, and \pithree) on various fisheye datasets, substantially outperforming strong baselines, including LoRA and calibration tokens.

\section{Related Work}
We review feed-forward 3D models, camera-aware geometry, projection-based reuse of perspective backbones, and lightweight adaptation.

\inparagraph{3D foundation models for image-based geometry.}
Classical reconstruction pipelines such as COLMAP~\citep{colmap}, as well as learning-based methods for individual matching or pose-estimation components~\citep{sarlin2021back,solonets2024analytical}, solve image-based geometry through modular stages. 
Recent 3D foundation models instead predict depth, pose, and coarse 3D structure in a feed-forward pass from only a few images~\citep{wang2024dust3r,duisterhof2025mastrsfm,wang2025vggt,wang2026pi3permutationequivariantvisualgeometry,lin2025depth}. 
Although these models differ in architecture and supervision, they are trained primarily on perspective imagery and therefore inherit pinhole-camera spatial priors.

\inparagraph{Fisheye, omnidirectional, and arbitrary-camera geometry.}
A separate line of work handles non-pinhole geometry by exposing the camera model to the network, including fisheye depth, omnidirectional depth, and universal camera-aware 3D estimation~\citep{mccraith2020monocular,xie2023omnividar,deng2025omnistereo,guo2025depth,zhao2025fisheyedepth,piccinelli2024unidepthuniversalmonocularmetric,piccinelli2025unik3d,hu2024metric3dv2}.
Other methods estimate or refine camera geometry explicitly, for example, through single-image calibration or radial-distortion averaging~\citep{tiradogarin2025anycalib,prada2025}.
Distortion-aware Transformers further modify tokenization, positional encoding, and dense decoding for wide-angle or fisheye images~\citep{athwale2023darswin,athwale2025darswin,yang2023sector}.
Camera-aware positional encodings are also closely related. PRoPE~\citep{li2025cameras} represents camera intrinsics and extrinsics as relative positional encodings for multi-view transformers, showing that camera geometry is an effective conditioning signal for feed-forward 3D tasks.
These works validate the importance of camera geometry, but require camera-aware training, new architectures, or a separate calibration/reconstruction procedure. Our goal is instead to adapt a frozen pinhole foundation model with a small online correction.
Contemporaneous works also adapt foundation models to fisheye input. Fisheye3R~\citep{duan2026fisheye3r} uses calibration tokens with masked attention, trained with supervision from synthetic fisheye distortions and optionally supervised fisheye data. FishRoPE~\citep{ahuja2026fishropeprojectiverotaryposition} introduces a fisheye-aware angular formulation of RoPE for supervised fisheye detection and BEV segmentation; its geometry-specific positional modification is restricted to RoPE, while the full model also trains LoRA modules and task-specific heads. In contrast, \ourMethod adapts already trained 3D foundation models for depth and pose by correcting both absolute and rotary positional components when present, keeps the backbone frozen, and fits the residuals self-supervised from a short real fisheye sequence.

\inparagraph{Projection-based adaptation to perspective models.}
Another strategy is to convert non-pinhole images into perspective-like views and reuse pinhole backbones. Examples include equirectangular or cubemap projection for depth estimation~\citep{guo2025depth,wang2024depthanywhere}, adaptive perspective slicing for VGGT-style models~\citep{yuan2026vggt360}, and cubemap alignment through graph optimization~\citep{jung2025rpg360}. 
These methods preserve the input distribution expected by the backbone, but duplicate computation across virtual views, require prediction fusion, and do not let the model represent the original fisheye geometry directly.

\inparagraph{Parameter-efficient and test-time adaptation.}
Parameter-efficient fine-tuning methods such as LoRA~\citep{hu2022lora}, adapter layers~\citep{houlsby2019parameter}, and SSF~\citep{lian2022scaling} update a small parameter subset while keeping the backbone fixed, but usually modify attention weights, MLP projections, prompts, or feature scales rather than the camera-dependent position-to-ray mapping. 
Calibration tokens~\citep{gangopadhyay2025extending} are more camera-aware and provide a strong fisheye baseline, but still express the correction through learned feature tokens. 
Test-time training methods optimize fast-adapting parameters online with self-supervised objectives~\citep{sun2020test,zhang2025testtimetrainingright,chen2025ttt3r,xie2026scal3rscalabletesttimetraining}. Unlike these methods, we do not train the backbone to support online optimization, and instead adapt only small positional corrections of an already frozen 3D model.

\section{Preliminaries}
\label{sec:prelim}

\inparagraph{Camera projection.}
We use a central camera model with projection $\kappa$ and inverse projection $\kappa^{-1}$, mapping between camera rays and image coordinates in $\Omega\subset\mathbb{R}^2$. For a pinhole camera,
\begin{equation}
    \kappa^{-1}(u,v)=\bigl((u-c_x)/f_x,\ (v-c_y)/f_y,\ 1\bigr)^\top,
\end{equation}
where $(f_x,f_y)$ are focal lengths and $(c_x,c_y)$ is the principal point.

Fisheye cameras change this pixel-to-ray map. In common radially symmetric models, a ray with angle \(\alpha\) to the optical axis lands at image radius \(\rho(\alpha)\) from the principal point. The function \(\rho(\alpha)\) can be a polynomial for the Kannala--Brandt (KB) model~\citep{kannala2006generic} or a rational function for the Enhanced Unified Camera Model (EUCM)~\citep{khomutenko2015enhanced}. Thus, unlike the pinhole case above, equal pixel displacements do not correspond to equal changes in viewing direction across the image.

\inparagraph{Patch tokens, positional embeddings and positional encoding.}
We study Transformer-based 3D foundation models, often initialized from DINOv2~\citep{oquab2024dinov2learningrobustvisual}. %
The Vision Transformer (ViT) splits  the image of size $H_{\mathrm{img}}\times W_{\mathrm{img}}$ into non-overlapping patches of size $p$, producing an $H\times W$ token grid with $H=H_{\mathrm{img}}/p$ and $W=W_{\mathrm{img}}/p$.
Each patch is embedded into a $C$-dimensional token.
During training (\eg on pinhole images), ViTs learn a table of positional tokens, $P_A\in\mathbb{R}^{H\times W\times C}$, added to the image tokens. 
If the input resolution changes from $H \times W$, ViTs usually adjust $P_A$ to the new grid size with bilinear interpolation.
However, the spatial structure of $P_A$ would still reflect the camera geometry observed at training time.

Some ViTs also use rotary positional encoding (RoPE)~\citep{heo2024rotary} in self-attention.
RoPE applies 2D rotations to pairs of query and key feature channels: For a token at position \(n\), channel pair $(2j, 2j + 1)$ is rotated by an angle proportional to \(n\omega_j \), with fixed frequencies \(\{\omega_j\}_j\).
This makes attention depend on relative token offsets rather than on absolute coordinates alone.
In image transformers, 2D axial RoPE applies the same construction separately along the vertical and horizontal grid coordinates.
We denote this rotary positional mechanism by \(P_R\).

\inparagraph{On pinhole bias.}
Trained on pinhole images, 3D foundation models cannot reliably handle images with a different camera geometry, such as fisheye images.
The geometric mismatch can be seen directly through the local backprojection Jacobian. Specifically, the Jacobian under the pinhole camera model,
\begin{equation}
    J_{\kappa^{-1}}=\partial\kappa^{-1}/\partial(u,v),
\end{equation}
is constant over the image.
Therefore, a one-pixel displacement induces the same local change of $1/f_{\{x,y\}}$ in viewing direction everywhere. 
By contrast, for a radially symmetric fisheye camera, the inverse projection can be written as
\begin{equation}\label{eq:fisheye_backproj}
\kappa^{-1}(u,v)=\bigl(g(u,v)(u-c_x)/f_x,\ g(u,v)(v-c_y)/f_y,\ 1\bigr)^\top,
\end{equation}
where $g$ is a non-linear function of image radius (\eg the pinhole model is a special case with $g\equiv1$).
Differentiating \cref{eq:fisheye_backproj} yields
\begin{equation}
J_{\kappa^{-1}}(u,v) = g(u,v)\begin{pmatrix}1/f_x & 0 \\ 0 & 1/f_y \\ 0 & 0\end{pmatrix} + \begin{pmatrix}(u-c_x)/f_x \\ (v-c_y)/f_y \\ 0\end{pmatrix}\bigl(\partial_u g,\ \partial_v g\bigr),
\label{eq:intro_jacobian}
\end{equation}
which now depends on the image coordinates $(u, v)$.
Thus, the same image-grid displacement has a different geometric meaning at different radii.
However, a positional encoding learned on pinhole images has only been exposed to the case of constant Jacobian, which leads to a pinhole bias.

\begin{figure}[t]
    \centering
    \begin{subfigure}[t]{0.24\linewidth}
        \centering
        \includegraphics[width=\linewidth]{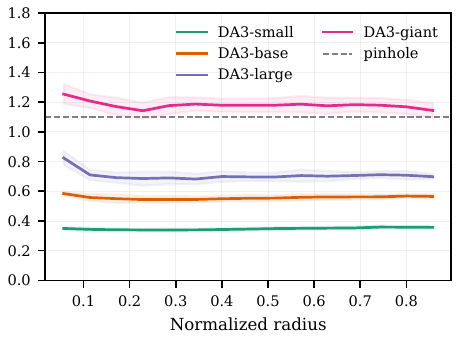}
        \caption{pretrained $\sigma_1$}
    \end{subfigure}\hfill
    \begin{subfigure}[t]{0.24\linewidth}
        \centering
        \includegraphics[width=\linewidth]{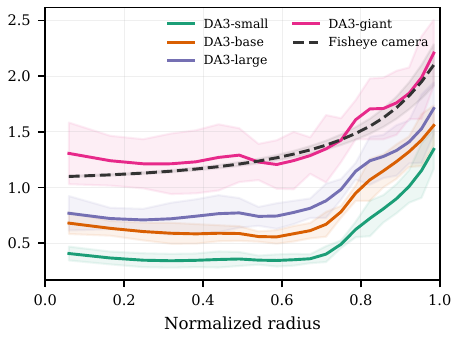}
        \caption{adapted $\sigma_1$}
    \end{subfigure}\hfill
    \begin{subfigure}[t]{0.24\linewidth}
        \centering
        \includegraphics[width=\linewidth]{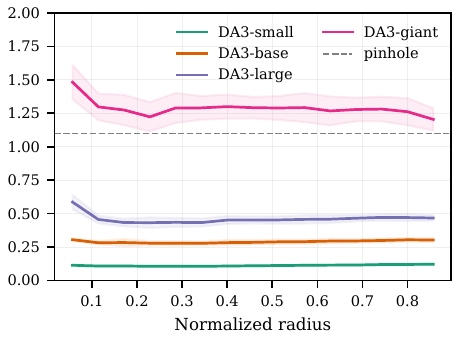}
        \caption{pretrained $\det$}
    \end{subfigure}\hfill
    \begin{subfigure}[t]{0.24\linewidth}
        \centering
        \includegraphics[width=\linewidth]{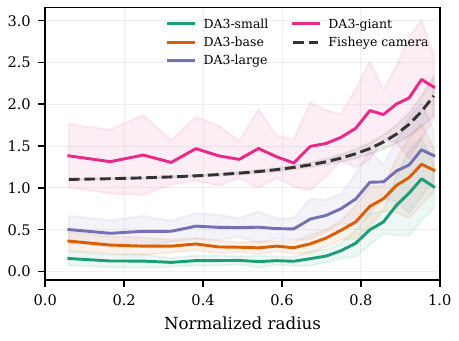}
        \caption{adapted $\det$}
    \end{subfigure}

 \caption{\textbf{Local Jacobian of the positional encoding in Depth Anything~3.}
    Pretrained positional embeddings have nearly radius-independent local geometry, consistent with a pinhole-like spatial prior (a, c).
    After adaptation on KITTI-360, the PE Jacobian becomes radius-dependent, matching the fisheye camera geometry more closely (b, d).}
    \label{fig:pe_jacobian_main}
\end{figure}

Using Depth Anything~3~\cite{lin2025depth}, we verify this bias by measuring the local Jacobian of the absolute positional embedding, \(J_{\mathrm{PE}}=\partial P_A/\partial(u,v)\in\mathbb{R}^{C\times2}\), and summarizing it by its largest singular value \(\sigma_1\) and local area element \(\sqrt{\det(J_{\mathrm{PE}}^\top J_{\mathrm{PE}})}\) in \cref{fig:pe_jacobian_main}.
These quantities capture the strongest local variation and local scale of the embedding. Across all frozen model sizes, both are nearly flat as a function of normalized radius \(\rho\) (\cf \cref{fig:pe_jacobian_main}a,b), matching the position-independent structure of a pinhole camera. After fitting our adapter on KITTI-360, the same curves bend toward the analytical fisheye reference (\cf \cref{fig:pe_jacobian_main}c,d).
This motivates \ourMethod: Finetuning the positional representations that carry the camera bias rather than generic feature processing.

\section{Method: \texorpdfstring{\ourMethod}{RayTun3R}}
Our goal is to adapt a pretrained pinhole 3D foundation model to fisheye input while keeping the pretrained model frozen. Fisheye optics change the mapping between token locations in the image and viewing rays, so the spatial structure learned during pinhole pretraining no longer matches the input camera.
Motivated by the PE Jacobian analysis, we therefore modify the components that encode how token locations in the image relate to geometry: Absolute positional embeddings, rotary positional encoding when present, prediction-grid coordinates, and patch tokenization.
The only learned parameters are small PE/RoPE lookup table corrections.
\cref{fig:da3_fisheye_failure} visualizes the main learned correction: The image content and backbone remain fixed, while the positional encoding is adapted to the fisheye camera geometry.

\subsection{Problem setup}
Let \(f_\theta\) be a pretrained 3D foundation model that predicts depth and pose from one or more images. We consider pinhole-trained backbones, such as Depth Anything~3, \pithree, and VGGT.
Given a fisheye image stream \(\{I_t\}\) and a known or estimated camera model $\kappa$, we learn a small set of adapter parameters \(\phi\) while keeping all backbone weights \(\theta\) fixed. The adapted model \(f_{\theta,\phi}\) should produce accurate depth and pose on fisheye input without retraining the original backbone.

The only trainable parts are attached to positional mechanisms. In particular, we adapt the absolute positional embedding \(P_A\) and, when present, the rotary positional mechanism \(P_R\) defined in \cref{sec:prelim}. 
All attention blocks, MLPs, prediction heads, and DPT weights remain frozen.

\subsection{Adapting positional embeddings}
The key difference between pinhole and fisheye geometry is radial, but real cameras and finite-resolution token grids may also have smaller orientation-dependent residuals.
We therefore parameterize the correction in polar coordinates around the calibrated principal point. For each patch at coordinate \(u, v\), let \(\rho_{u,v} \in [0,1]\) be its normalized radius and \(\theta_{u,v} \in [0,2\pi)\) its angle, both computed \wrt $(c_x,c_y)$. 
We discretize radius into \(N_r\) bins and angle into \(N_\theta\) bins, and learn a radial update \(t_r(\rho)\) together with an angular update \(\delta_\theta(\theta)\).
Both updates are implemented as lookup tables and evaluated at continuous \(\rho_{u,v}\) and \(\theta_{u,v}\) by linear interpolation over the radial and angular bins.
The corrected positional embedding is
\begin{equation}\label{eq:pe_correction}
P'(u,v)=P_A(u,v)+t_r(\rho_{u,v})+\rho_{u,v}\,\delta_\theta(\theta_{u,v}).
\end{equation}
The factor \(\rho_{u,v}\) suppresses the angular term at the image center, where the angle is ill-defined, and lets it grow toward the periphery.
We compute it as a patch radius, normalized to \([0,1]\) by the grid boundary.
All adapter parameters are initialized to zero, so adapter training starts exactly from the pretrained positional table.

\inparagraph{Rotary positional encoding.}
For backbones that use RoPE, we learn a radial correction to the rotary angles. 
If \(\omega(u,v)\) is the original rotary angle at patch \((u, v)\), we add a correction from a learnable radial lookup table (shared across RoPE frequencies) with one parameter per bin:
\begin{equation}\label{eq:ro_correction}
\omega'(u,v)=\omega(u,v)+\Delta_r(\rho_{u,v}).
\end{equation}

\inparagraph{Prediction-grid coordinates.}
Some depth heads use a Dense Prediction Transformer (DPT)-style~\citep{ranftl2021visiontransformersdenseprediction} head with a 2D sinusoidal prediction grid.
We replace the regular prediction-grid coordinates with camera-aware coordinates obtained by undistorting each fisheye grid location through the calibrated fisheye-to-pinhole map. This does not change any model weights.

\inparagraph{Patch tokenization.}
Patches outside the valid fisheye lens circle are replaced by the mean valid token, avoiding artifacts from the invalid black region (visible as the dark boundary in \cref{fig:da3_fisheye_failure}a). We also locally undistort each patch before tokenization. Since fisheye distortion is position-dependent, especially near the boundary, each patch is resampled using the local linearization of the fisheye-to-pinhole map at its center following~\citet{qin2012tracking}. This parameter-free step makes each tokenized patch closer to a pinhole crop from the same viewing direction.

These learned and parameter-free changes add negligible inference overhead but improve stability and accuracy in practice.

\subsection{Adapter training}
We fit the adapter online on a short temporal segment of fisheye frames. 
Training uses windows of three-frames \(\{I_i\}_{i=1}^3\) sampled from this segment. 
For each window,  the adapted model \(f_{\theta,\phi}\) predicts a depth map \(D_i\) and camera pose \(T_i\) for each image. A predicted depth value defines 
\begin{equation}
X_i(u,v)=D_i(u,v)\,\kappa^{-1}(u,v),
\end{equation}
a 3D point at pixel \((u,v)\), expressed in the coordinate frame of image \(I_i\).

For each ordered image pair \((I_i,I_j)\), let \(m_{ij}(u,v)\) be the matched pixel in image \(I_j\) for pixel \((u, v)\) in image \(I_i\), and let \(w_{ij}(u,v)\) be its confidence.
To estimate the matching pixels, we use the correspondences produced by an off-the-shelf network, UFM~\citep{zhang2025ufm}, and their confidence scores.
We penalize the reprojection error after transforming the predicted 3D point from \(I_i\) to \(I_j\):
\begin{equation}
L_{\mathrm{reproj}}(i,j)=
\frac{1}{|\Omega|}
\sum_{(u,v)\in\Omega}
w_{ij}(u,v)\,
\left\|
\kappa\!\left(T_jT_i^{-1}X_i(u,v)\right)-m_{ij}(u,v)
\right\|_1,
\end{equation}
where $\Omega$ denotes the set of pixels inside the valid fisheye disc.

The reprojection loss couples depth and pose, but pose optimization can be unstable early in adaptation.
Therefore, we add a fixed geometric pose target. 
From the same UFM matches, we estimate a relative pose \(\tilde{T}_{ij}=(\tilde{R}_{ij},\tilde{t}_{ij})\) once using MAGSAC++~\citep{barath2020magsac++}. We then penalize the angular difference between this estimate and the model-predicted relative pose \(T_jT_i^{-1}=(\hat{R}_{ij},\hat{t}_{ij})\):
\begin{equation}
L_{\mathrm{pose}}(i,j)=
\arccos\!\left(
\frac{\mathrm{tr}\!\left(\tilde{R}_{ij}^{\top}\hat{R}_{ij}\right)-1}{2}
\right)
+
\arccos\!\left(
\frac{\langle \tilde{t}_{ij},\hat{t}_{ij}\rangle}
{\|\tilde{t}_{ij}\|\,\|\hat{t}_{ij}\|}
\right).
\end{equation}
Since \(\tilde{T}_{ij}\) is computed once from correspondences produced by an external matcher before adaptation, the adapter cannot influence its own pose pseudo-label.
This fixed pose target stabilizes relative motion early in optimization, while \(L_{\mathrm{reproj}}\) enforces depth consistency through the predicted geometry.

We further regularize depth with the standard edge-aware smoothness prior from self-supervised monocular depth estimation~\citep{godard2017unsupervised}:
\begin{equation}
L_{\mathrm{smooth}}(i)=
\frac{1}{|\Omega|}
\sum_{(u,v)\in\Omega}
e^{-|\partial_x I_i(u,v)|}\,|\partial_x D_i^\ast(u,v)| + e^{-|\partial_y I_i(u,v)|}\,|\partial_y D_i^\ast(u,v)|,
\end{equation}
where \(D_i^\ast = D_i / \mathrm{mean}(D_i)\).

Finally, we keep the positional correction small and smooth.
We regularize the adapted positional table \(P' \in \mathbb{R}^{H_p \times W_p \times C}\) toward the initial table \(P_A\):
\begin{equation}
\mathcal{L}_{\mathrm{L2}}
=
\frac{1}{H_pW_p}\sum_{x,y}
\,\|P'(x,y)-P_A(x,y)\|_2^2,
\end{equation}
where \(x,y\) index token-grid locations. We further constrain neighboring positional corrections to vary smoothly using a total-variation penalty:
\begin{equation}
\mathcal{L}_{\mathrm{TV}}
=
\frac{1}{H_pW_p}\sum_{x,y}
\Bigl(
\|P'(x+1,y)-P'(x,y)\|_2^2
+
\|P'(x,y+1)-P'(x,y)\|_2^2
\Bigr),
\end{equation}
where differences are taken only for valid neighbors. The total loss is
\begin{equation}
\label{eq:training_loss}
L=
L_{\mathrm{reproj}}
+
w_{\mathrm{smooth}}L_{\mathrm{smooth}}
+
w_{\mathrm{L2}}\mathcal{L}_{\mathrm{L2}}
+
w_{\mathrm{TV}}\mathcal{L}_{\mathrm{TV}}
+
w_{\mathrm{pose}}\mathcal{L}_{\mathrm{pose}}.
\end{equation}

We use $ w_{\mathrm{pose}} = 1$, $w_{\mathrm{smooth}} = 10$, $w_{\mathrm{L2}} = 2$, and $w_{\mathrm{TV}} = 20$ in all experiments. \cref{sec:ablation_loss} ablates the contribution of the pose and PE regularization terms.

We do not use a photometric warping loss. Direct photometric alignment can be weak in low-texture regions and sensitive to initialization. The initial depth and pose can be far from optimum due to the model's pinhole bias. Following recent geometry and pose predictors such as AnyCam~\citep{wimbauer2025anycamlearningrecovercamera}, we instead rely on UFM correspondences and optimize geometric consistency.

In summary, the only learned parameters are small radial/angular lookup tables for positional encodings; all backbone weights and prediction heads remain frozen, while the DPT-grid and patch-tokenization changes are parameter-free.

\inparagraph{Implementation details}
Unless stated otherwise, images are resized to a maximum patch-aligned resolution of \(504\times504\). We fit each adapter using batches of three-frame windows with Adam, learning rate \(1\times10^{-3}\), and gradient clipping at norm 1.0.
The PE adapter uses 20 radial bins and 8 angular bins; RoPE backbones additionally use 20 radial RoPE bins. All backbone weights remain frozen, and all residual adapter parameters are initialized to zero. After adaptation, inference requires a single forward pass per fisheye frame.

\section{Experiments}
\label{sec:experiments}

\inparagraph{Datasets.}
Fisheye datasets with reliable pose or depth ground truth remain less common than standard pinhole reconstruction benchmarks.
We evaluate on five datasets covering outdoor driving (KITTI-360~\citep{liao2022kitti360noveldatasetbenchmarks}, \(185^\circ\) FOV), handheld dual-fisheye video (TUM-VI~\citep{Schubert_2018}, \(195^\circ\)),
\begin{wrapfigure}[11]{r}{0.43\textwidth}
    \centering
    \vspace{-0.2em}
    \begin{subfigure}[t]{0.47\linewidth}
        \centering
        \includegraphics[width=\linewidth]{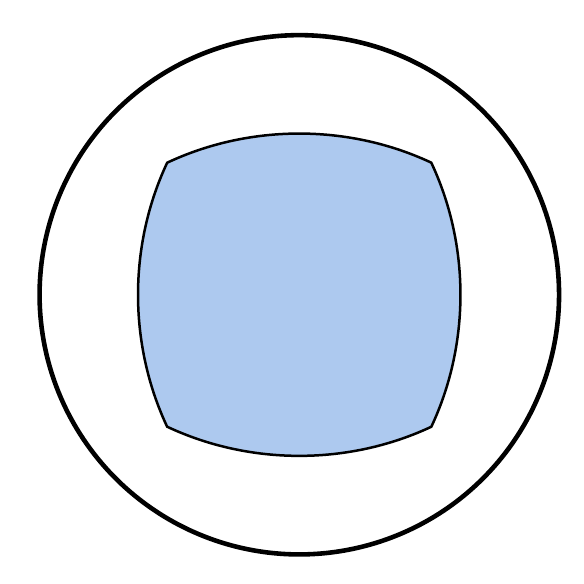}
        \caption{Center-PH}\label{fig:pinhole_coverage_a}
    \end{subfigure}
    \hfill
    \begin{subfigure}[t]{0.47\linewidth}
        \centering
        \includegraphics[width=\linewidth]{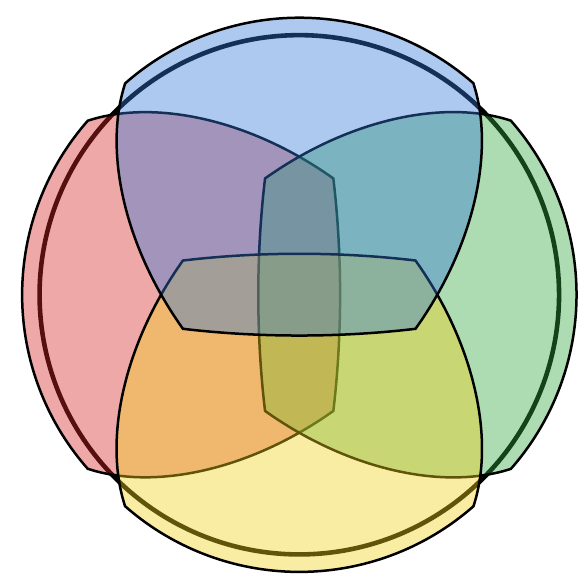}
        \caption{Multi-PH}\label{fig:pinhole_coverage_b}
    \end{subfigure}
    \caption{\textbf{Adaptation-free baselines.}}
    \label{fig:pinhole_coverage}
    \vspace{-1em}
\end{wrapfigure}
posed DSLR fisheye scans (ScanNet++~\citep{yeshwanth2023scannethighfidelitydataset3d}, \(115^\circ\)), multi-camera captures (ETH3D~\citep{schops2017multi}, \(110^\circ\)), and dual-fisheye indoor/outdoor scenes (FIORD~\citep{gunes2025fiordfisheyeindooroutdoordataset}, \(200^\circ\)).
All datasets provide reference poses; ETH3D and ScanNet++ also provide dense depth.

\inparagraph{Baselines.}
We compare against representative baselines from the two most relevant strategies for reusing pinhole-trained models on fisheye input: input-space projection and lightweight model adaptation.
Center-PH undistorts each fisheye image into one forward-looking \(110^\circ\) virtual pinhole crop (\cref{fig:pinhole_coverage_a}); we use \(110^\circ\) to avoid severe center compression and peripheral stretching from wider rectilinear projections.
Multi-PH fuses predictions from Center-PH and four additional virtual pinhole views back into the fisheye frame (\cref{fig:pinhole_coverage_b}), improving coverage at roughly proportional extra cost.

For adaptation-based baselines, we include CalTok~\citep{gangopadhyay2025extending}, the closest baseline to our setting because it adapts mostly frozen depth networks to fisheye cameras. CalTok adds \(t\in\{4,8\}\) learned auxiliary tokens to attention layers. We also include LoRA~\citep{hu2022lora} as a strong generic PEFT baseline, updating QKV adapters with \(r\in\{4,8,16\}\) and \(\alpha\in\{8,16,32\}\).
For the main comparison, we report the strongest ETH3D setting for each baseline: CalTok with \(t=4\) and LoRA with \(r=8,\alpha=16\) (see \cref{tab:depth_regime}, right).
Unless specified otherwise, experiments use DA3-Small~\citep{lin2025depth}; we also evaluate larger DA3 variants, \pithree~\citep{wang2026pi3permutationequivariantvisualgeometry}, and VGGT~\citep{wang2025vggt}.

\inparagraph{Error metrics.}
For camera pose accuracy, we report angular rotation error $R^\circ$ and translation-direction error $t^\circ$, both in degrees.
Since ground-truth dense depth is not available for all fisheye datasets, we use a reprojection-based depth metric \emph{with ground-truth pose} for the main comparison, denoted by $d_{\mathrm{reproj}}$.
When reliable ground-truth depth is available, we report scale-aligned depth error, $\text{AbsRel}$, following~\citet{eigen2014depth}, and the threshold accuracy \(\delta_{1.25}\), which is the fraction of pixels whose predicted depth is within a factor \(1.25\) of the ground truth. \cref{sec:err_metrics} provides details on metric computation.

{
\newcolumntype{Y}{>{\centering\arraybackslash}X}

\begin{table}[t]
\centering
\caption{\textbf{Pose and depth error across five datasets using DA3-Small.} Each entry reports a triple $R^{\circ}, t^{\circ}, d_{\text{reproj}}$, respectively.
\textbf{Bold} and \uline{underline} denote the best and second-best results, respectively.
\ourMethod consistently offers low pose and depth error for the complete fisheye image (in contrast to Center-PH) and without extra inference-time costs (in contrast to Multi-PH). 
}
\label{tab:cross_dataset}
\scriptsize
\setlength{\tabcolsep}{0.5pt}

\newcommand{\triple}[3]{%
  \makebox[2.5em][r]{#1}%
  \hspace{0.25em}%
  \makebox[2.4em][r]{#2}%
  \hspace{0.25em}%
  \makebox[2.4em][r]{#3}%
}

\begin{tabularx}{\linewidth}{@{}lYYYYYY@{}}
\toprule
Dataset & Vanilla & Center-PH & Multi-PH & LoRA & CalTok & {\textbf{\ourMethod (ours)}} \\
\midrule

ETH3D
& \triple{8.59}{15.16}{15.98}
& \triple{3.46}{13.70}{10.92}
& \triple{3.31}{13.68}{13.48}
& \triple{\uline{2.18}}{\uline{10.74}}{\uline{9.02}}
& \triple{2.48}{13.21}{11.94}
& \triple{\textbf{0.70}}{\textbf{4.48}}{\textbf{5.82}} \\

KITTI-360
& \triple{1.69}{12.81}{11.64}
& \triple{\textbf{0.79}}{\uline{4.17}}{\textbf{3.10}}
& \triple{1.71}{9.75}{4.72}
& \triple{1.37}{8.49}{5.56}
& \triple{1.66}{10.05}{5.83}
& \triple{\uline{0.84}}{\textbf{2.92}}{\uline{3.88}} \\

TUM-VI
& \triple{10.41}{23.23}{57.01}
& \triple{3.33}{29.24}{\textbf{3.22}}
& \triple{\uline{2.99}}{25.60}{4.92}
& \triple{3.38}{\uline{13.63}}{3.83}
& \triple{3.84}{16.17}{9.61}
& \triple{\textbf{2.41}}{\textbf{13.23}}{\uline{3.81}} \\

ScanNet++
& \triple{10.21}{30.26}{23.82}
& \triple{3.27}{22.77}{\uline{2.21}}
& \triple{\uline{1.66}}{\uline{10.43}}{\textbf{1.63}}
& \triple{3.68}{17.66}{4.98}
& \triple{4.51}{23.20}{7.02}
& \triple{\textbf{1.11}}{\textbf{5.78}}{4.16} \\

FIORD
& \triple{18.20}{29.50}{75.30}
& \triple{6.92}{23.40}{\textbf{7.20}}
& \triple{\uline{6.30}}{18.90}{15.60}
& \triple{7.75}{\uline{12.20}}{12.10}
& \triple{20.40}{22.20}{25.20}
& \triple{\textbf{4.10}}{\textbf{5.40}}{\uline{9.00}} \\

\bottomrule
\end{tabularx}
\end{table}

}

\inparagraph{Evaluation protocol.}
For each sequence, we build a short adaptation set of 30 three-frame windows, filtering out nearly static windows with average optical-flow displacement below \(2\) pixels. We fit the adapter on this set and evaluate on the full sequence.

We evaluate relative poses on consecutive image pairs.
Note that short baselines can amplify translation-direction errors ($t^\circ$) even when reprojection error  ($d_{\mathrm{reproj}}$) remains moderate.
All methods use the same pair sampling.
Unless stated otherwise, all methods use the calibration provided with each dataset; \cref{sec:anycalib} evaluates sensitivity to predicted calibration.

\begin{figure}[t]
\centering
\setlength{\tabcolsep}{1pt}
\begin{subfigure}[t]{0.49\linewidth}
\centering
\begin{tabular}{@{}ccc@{}}
\makebox[0.32\linewidth][c]{\includegraphics[width=0.32\linewidth]{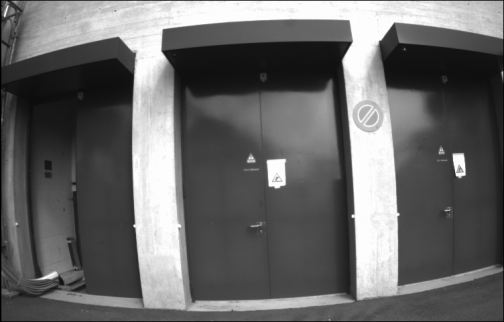}} &
\makebox[0.32\linewidth][c]{\includegraphics[width=0.32\linewidth]{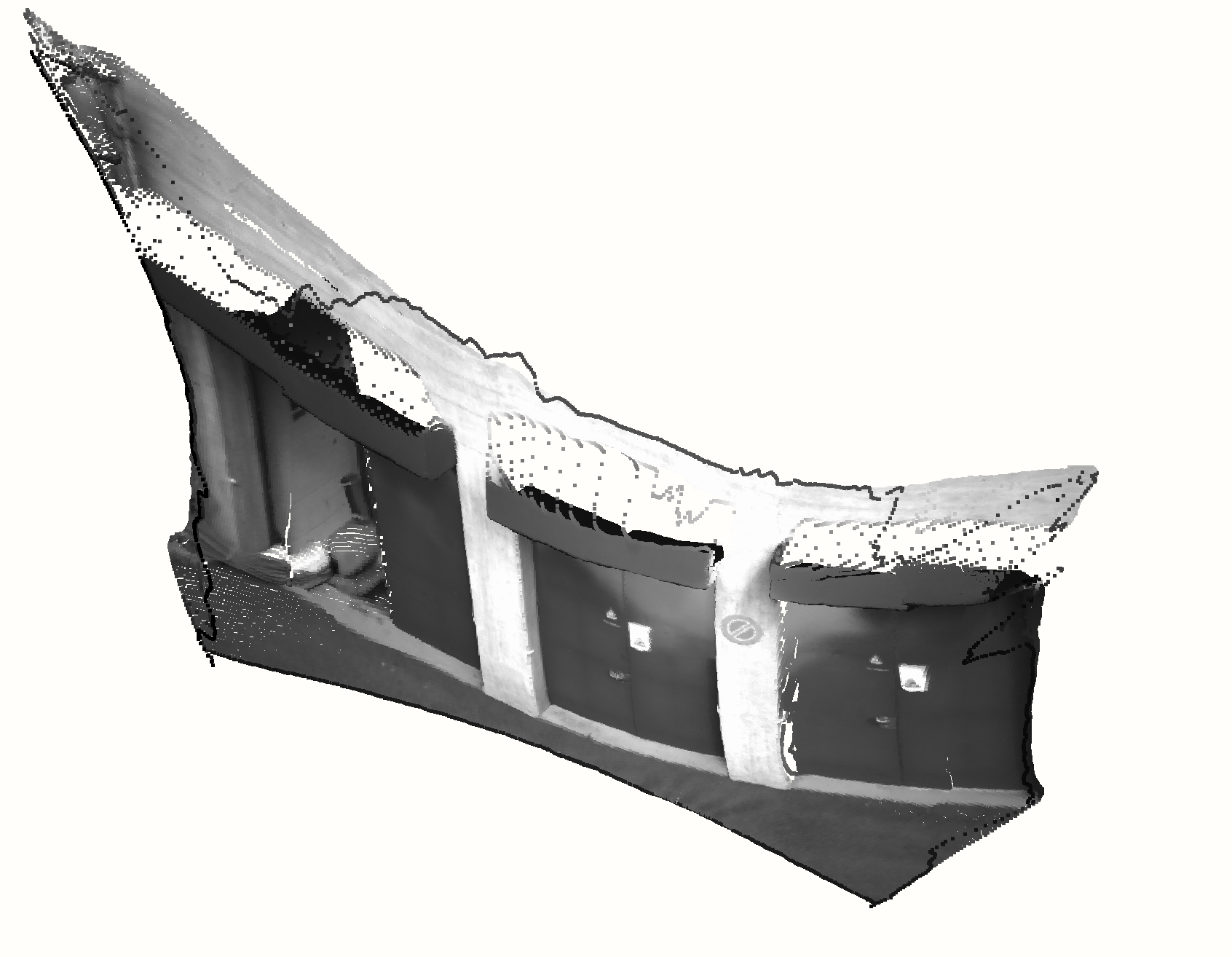}} &
\makebox[0.32\linewidth][c]{\includegraphics[width=0.32\linewidth]{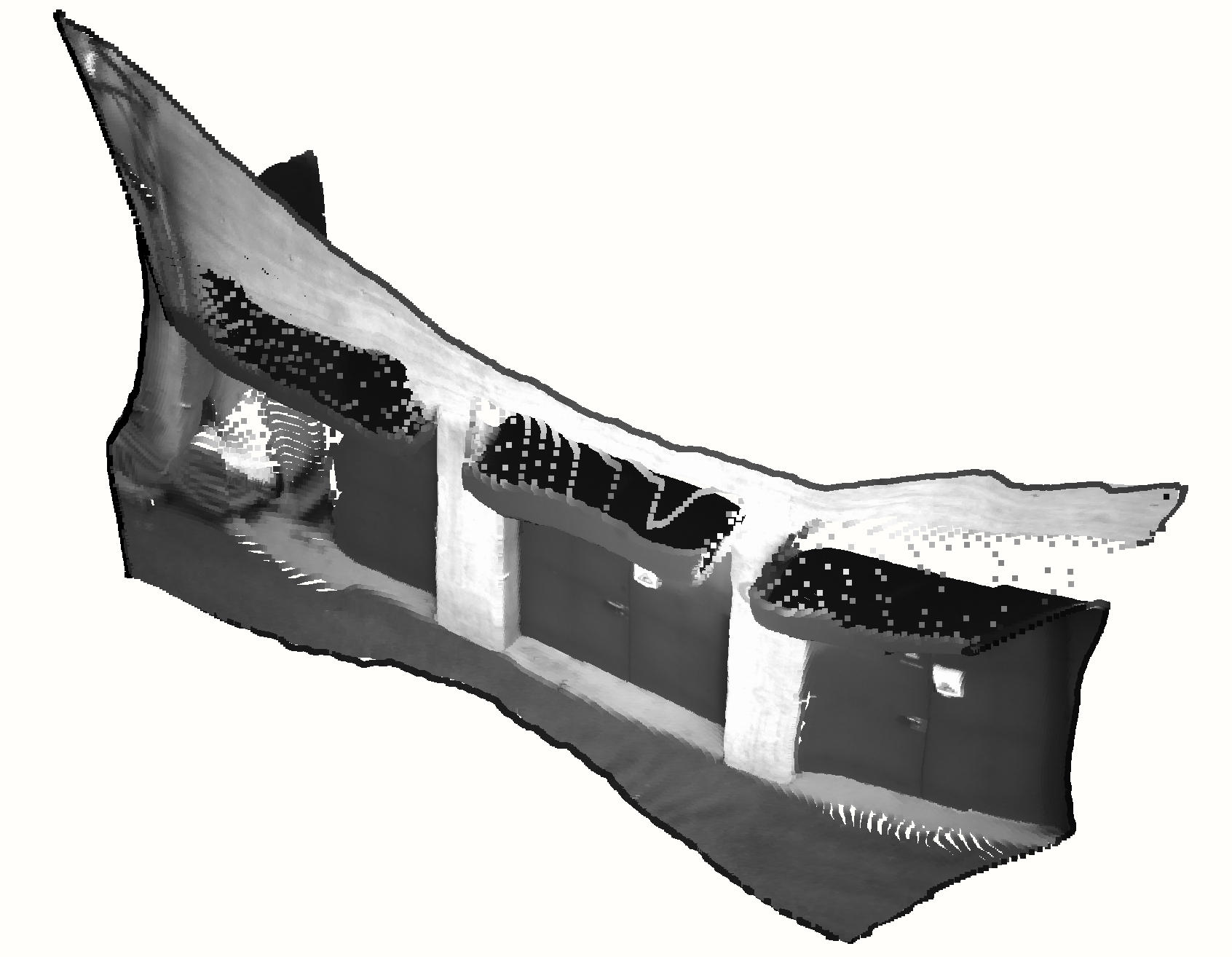}}
\end{tabular}
\end{subfigure}
\hfill
\begin{subfigure}[t]{0.49\linewidth}
\centering
\begin{tabular}{@{}ccc@{}}
\makebox[0.32\linewidth][c]{\includegraphics[width=0.32\linewidth]{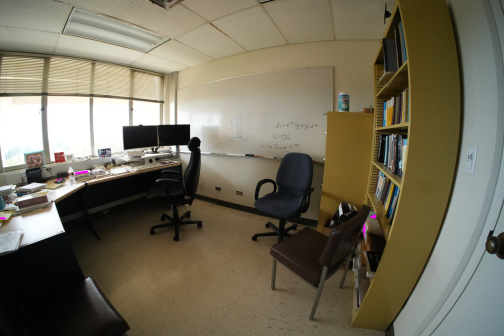}} &
\makebox[0.32\linewidth][c]{\includegraphics[width=0.32\linewidth]{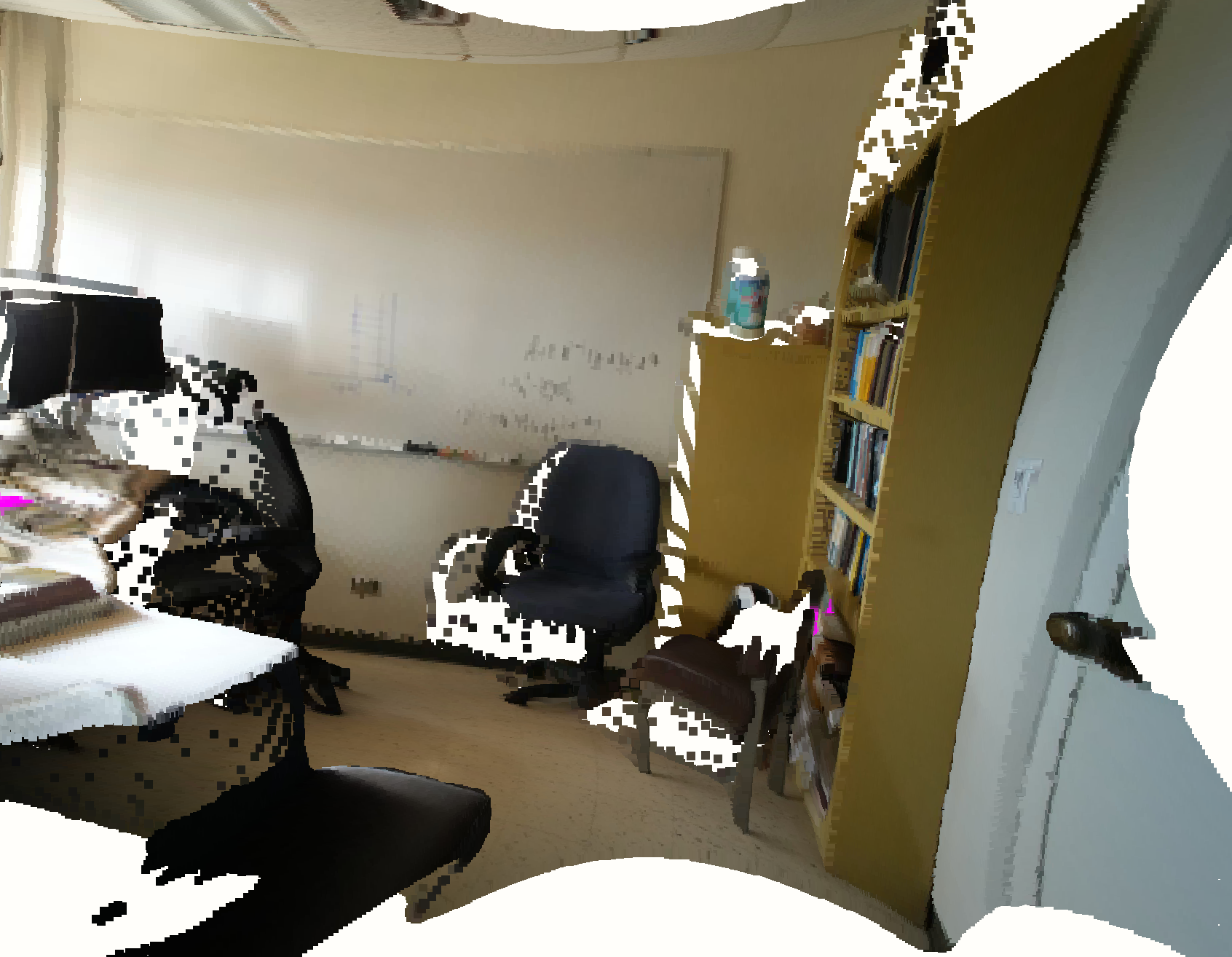}} &
\makebox[0.32\linewidth][c]{\includegraphics[width=0.32\linewidth]{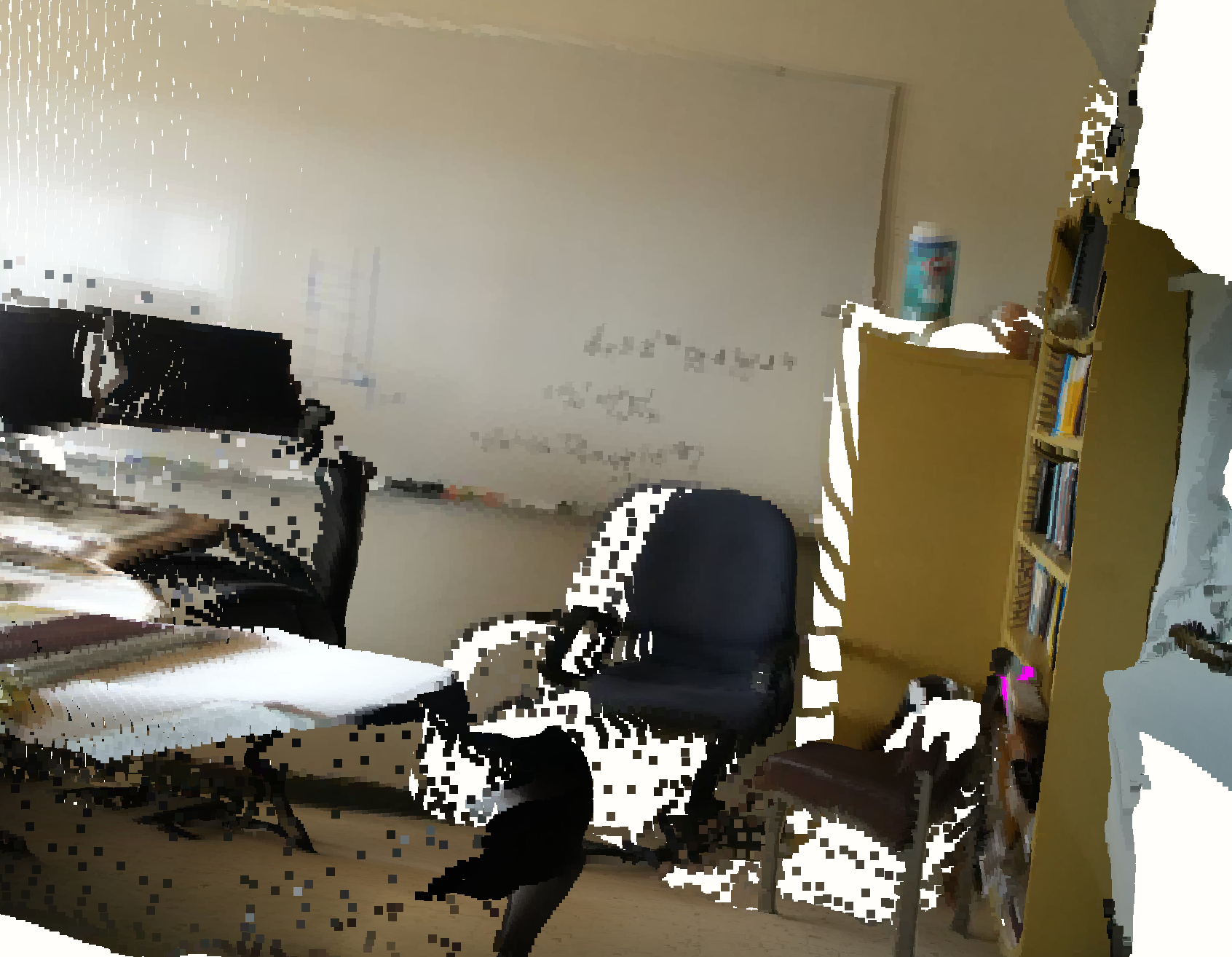}}
\end{tabular}
\end{subfigure}

\vspace{1mm}

\begin{subfigure}[t]{0.49\linewidth}
\centering
\begin{tabular}{@{}ccc@{}}
\makebox[0.32\linewidth][c]{\includegraphics[width=0.32\linewidth]{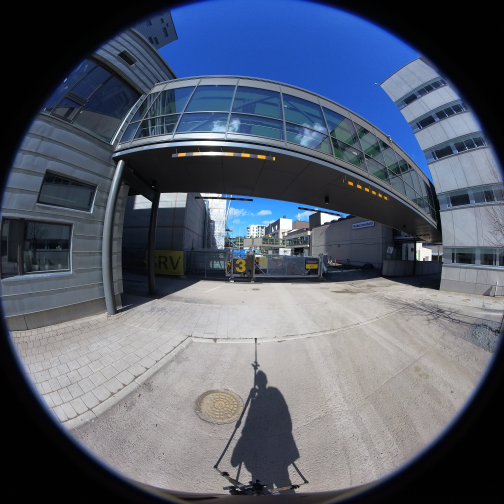}} &
\makebox[0.32\linewidth][c]{\includegraphics[width=0.32\linewidth]{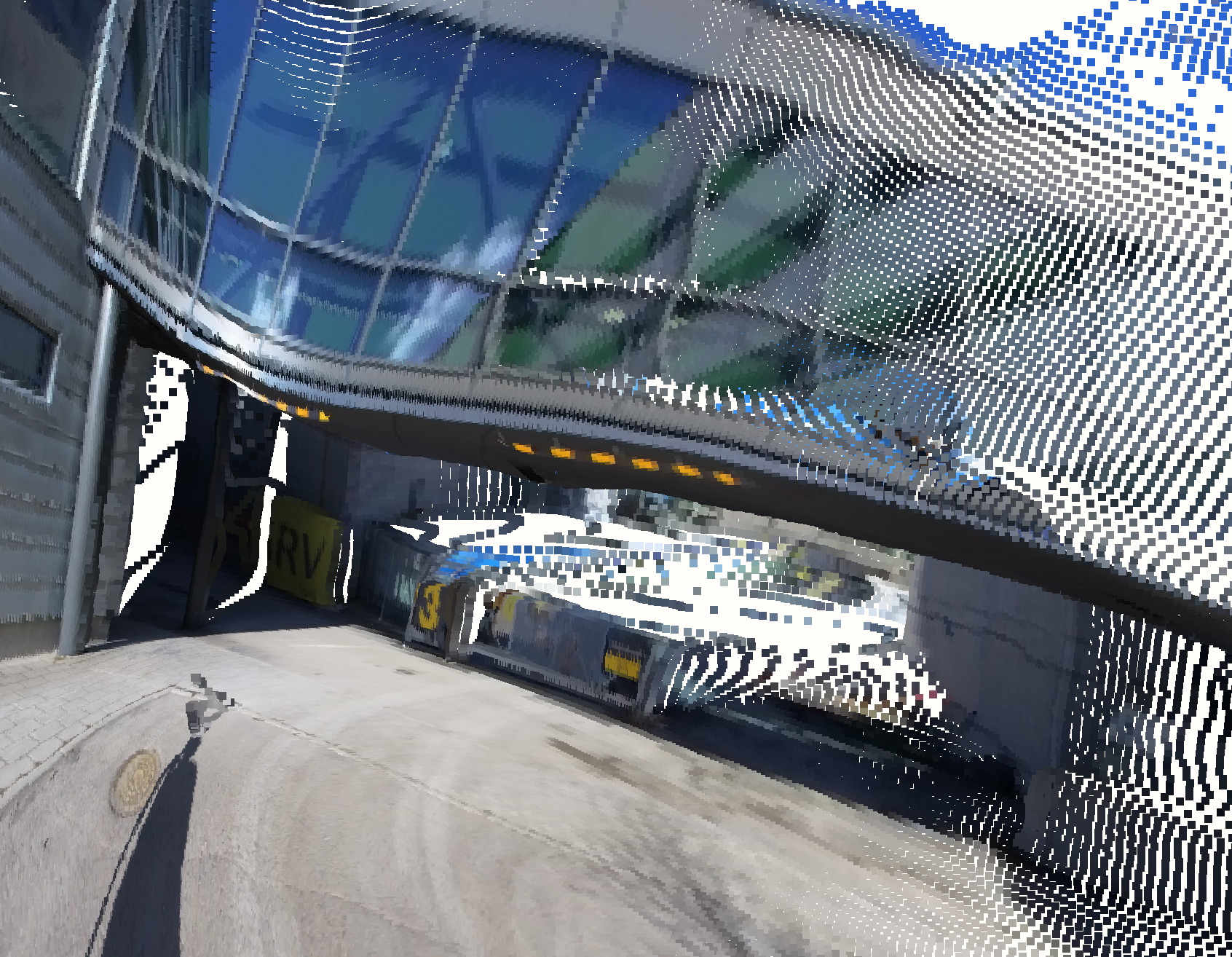}} &
\makebox[0.32\linewidth][c]{\includegraphics[width=0.32\linewidth]{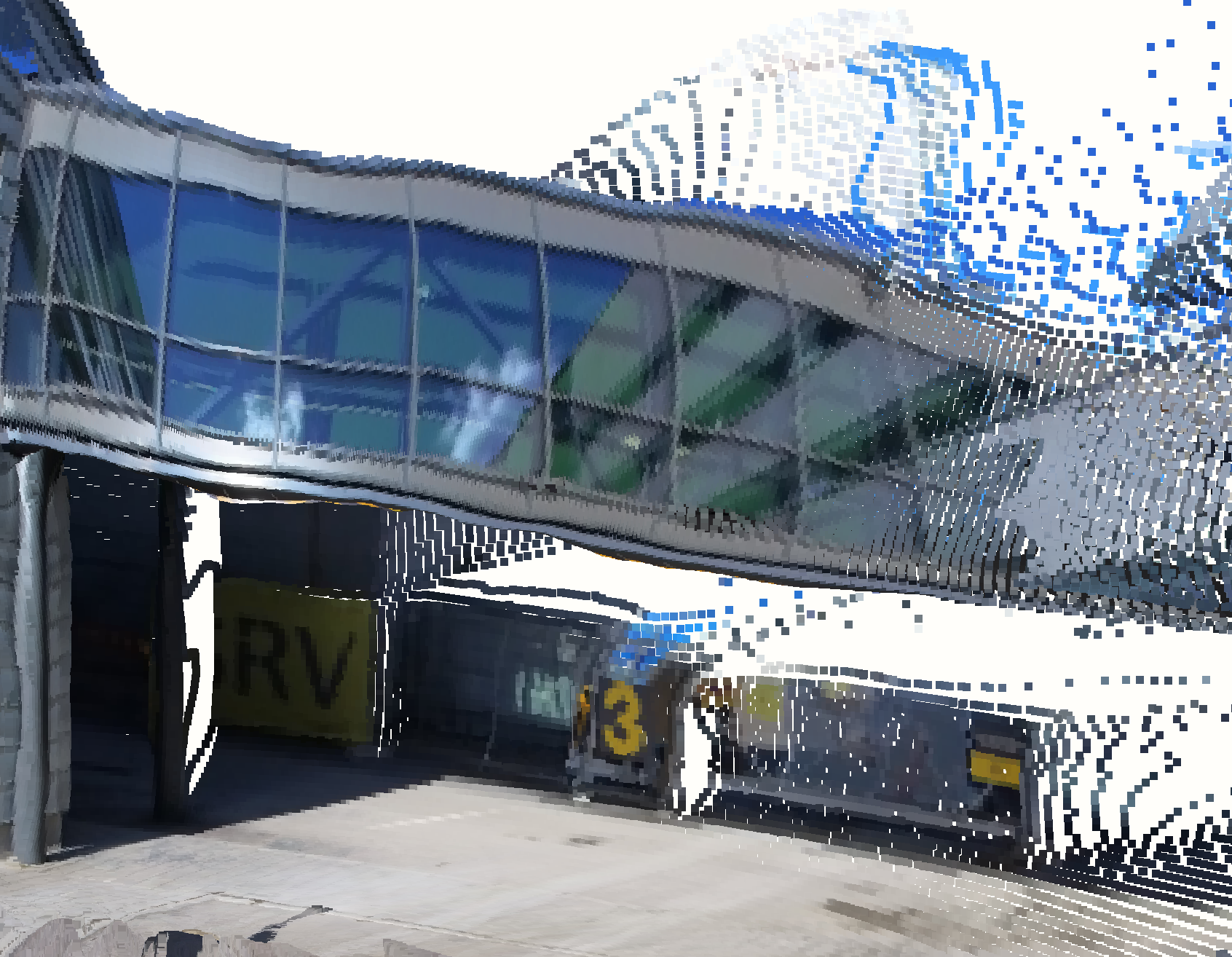}} \\
[-0.2em]
\makebox[0.32\linewidth][c]{\scriptsize Input} &
\makebox[0.32\linewidth][c]{\scriptsize Depth Anything 3} &
\makebox[0.32\linewidth][c]{\scriptsize \textbf{w/ \ourMethod (ours)}}
\end{tabular}
\end{subfigure}
\hfill
\begin{subfigure}[t]{0.49\linewidth}
\centering
\begin{tabular}{@{}ccc@{}}
\makebox[0.32\linewidth][c]{\includegraphics[width=0.32\linewidth]{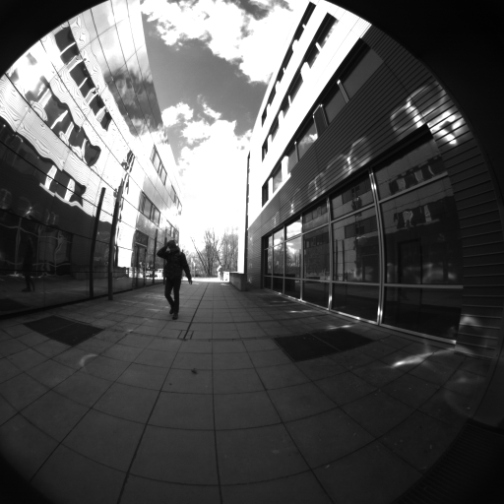}} &
\makebox[0.32\linewidth][c]{\includegraphics[width=0.32\linewidth]{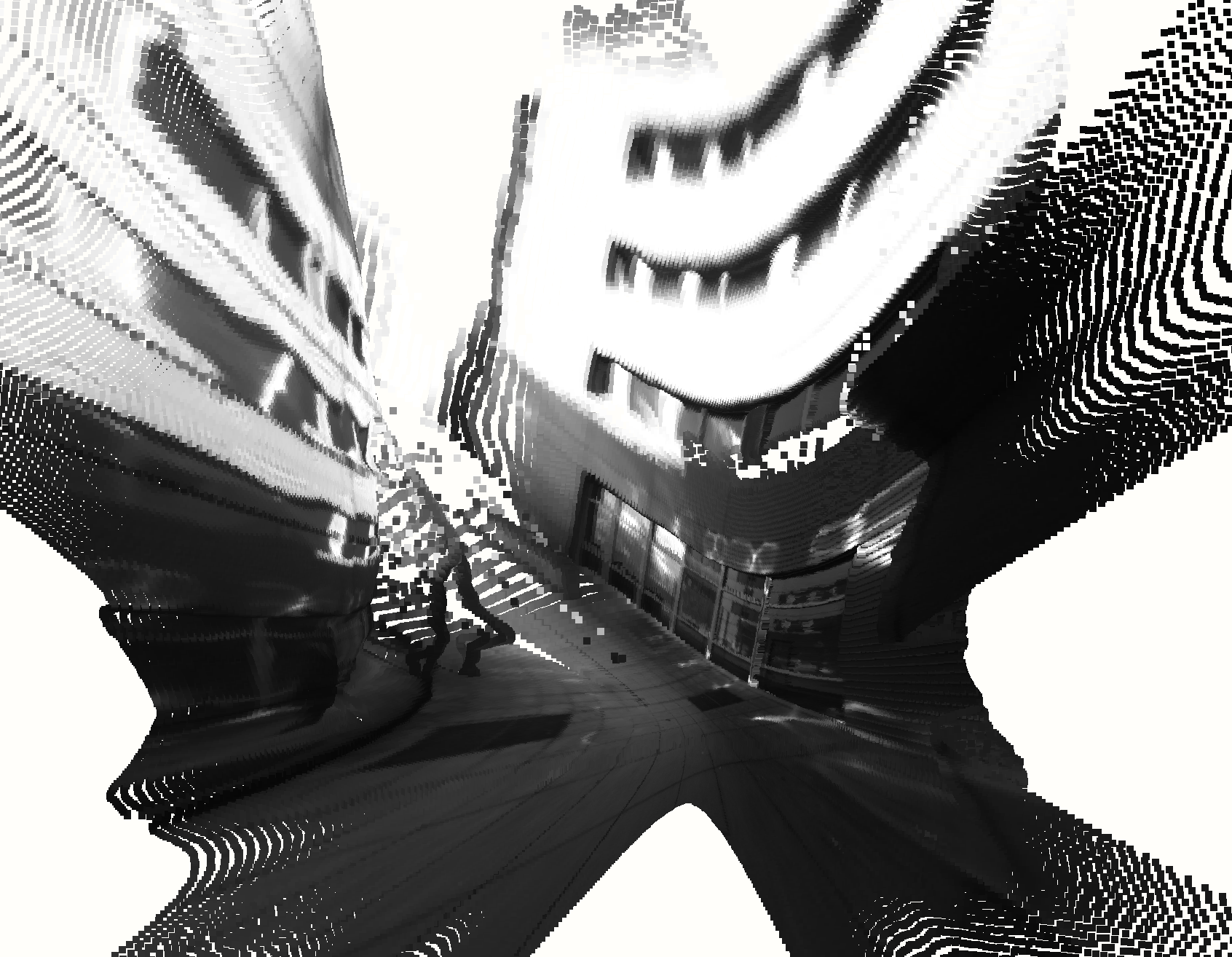}} &
\makebox[0.32\linewidth][c]{\includegraphics[width=0.32\linewidth]{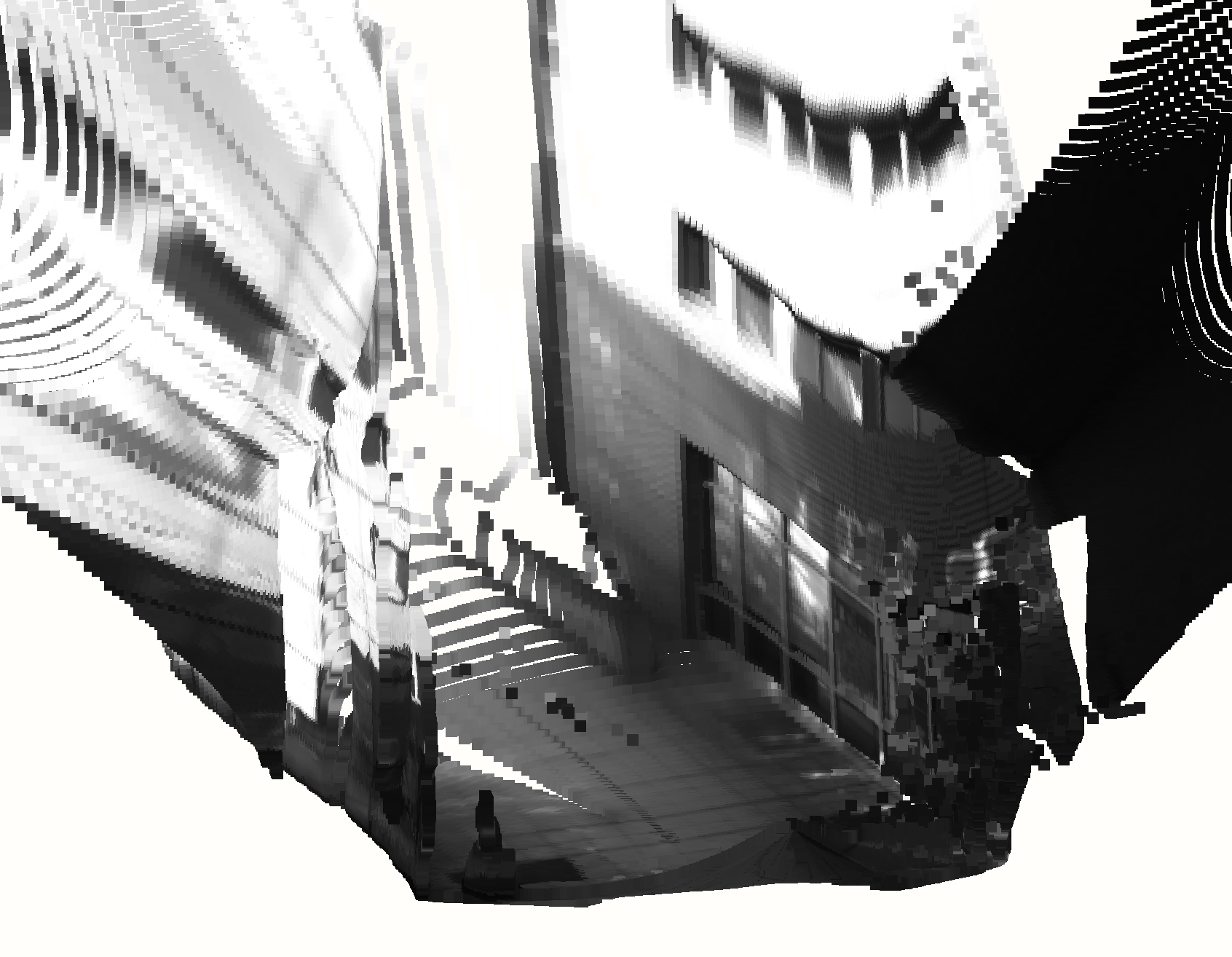}} \\
[-0.2em]
\makebox[0.32\linewidth][c]{\scriptsize Input} &
\makebox[0.32\linewidth][c]{\scriptsize Depth Anything 3} &
\makebox[0.32\linewidth][c]{\scriptsize \textbf{w/ \ourMethod (ours)}}
\end{tabular}
\end{subfigure}

\caption{\textbf{Qualitative reconstructions.}
Across datasets, the frozen model produces distorted or inconsistent geometry, while \ourMethod recovers more coherent structure after online adaptation.}
\label{fig:qualitative_main}
\end{figure}

\subsection{Main results}
\label{sec:main_results}

In \cref{tab:cross_dataset}, \ourMethod gives the lowest translation-direction error on all datasets and the lowest rotation error on ETH3D, TUM-VI, ScanNet++, and FIORD.
Compared with CalTok, the closest fisheye-specific adaptation baseline, and LoRA, the strongest generic PEFT baseline, \ourMethod achieves substantially lower pose error and lower or comparable reprojection. 
This supports our view that camera adaptation is better handled by directly correcting camera-dependent positional components than by adapting generic feature transformations. Such transformations do not explicitly separate camera geometry from scene appearance.
Compared with projection baselines, \ourMethod keeps the full fisheye field of view and single-view inference cost; Center-PH remains strong for depth on some moderate-FOV datasets, while Multi-PH improves coverage at a higher cost.
\cref{fig:qualitative_main} visually shows that, on fisheye inputs, adaptation produces more coherent geometry than the vanilla DA3 model.

\cref{tab:model_family_comparison} extends the comparison to \pithree and VGGT.
Across both backbones, \ourMethod gives the lowest translation error on every sequence and consistently ranks among the best methods for rotation.
It also improves substantially over CalTok and LoRA, indicating that the effect is not specific to DA3-Small.
\cref{sec:all_backbones_eth3d} reports additional results across DA3 model variants, \pithree, and VGGT.

Together, \cref{tab:cross_dataset,tab:model_family_comparison} support the core idea behind \ourMethod: Correcting camera-dependent positional components improves fisheye geometry across datasets and frozen 3D backbones.

{

\newcolumntype{Y}{>{\centering\arraybackslash}X}

\newcommand{\triple}[3]{%
  \makebox[2.5em][r]{#1}%
  \hspace{0.5em}%
  \makebox[2.2em][r]{#2}%
  \hspace{0.5em}%
  \makebox[1.8em][r]{#3}%
}

\begin{table}[t]
\centering
\caption{\textbf{Model family comparison under ground-truth calibration.} Each entry reports $R^{\circ}, t^{\circ}, d_{\mathrm{reproj}}$, respectively; lower is better. \textbf{Bold} and \uline{underline} denote the best and second-best results, respectively.}
\label{tab:model_family_comparison}

\scriptsize
\setlength{\tabcolsep}{2.5pt}

\begin{tabularx}{\linewidth}{@{}llYYYYY@{}}
\toprule
Backbone & Method & ETH3D terrains & KITTI 0009 & TUM-VI room6 & ScanNet++ 3f15 & FIORD Kitchen \\
\midrule

\multirow{5}{*}{\pithree~\cite{wang2026pi3permutationequivariantvisualgeometry}}
& LoRA
& \triple{1.40}{4.9}{1.4}
& \triple{2.22}{\uline{3.4}}{2.7}
& \triple{9.01}{22.4}{6.8}
& \triple{3.11}{\uline{16.1}}{\uline{2.1}}
& \triple{4.59}{4.62}{\textbf{5.5}}
\\

& CalTok
& \triple{1.49}{7.5}{1.3}
& \triple{0.93}{11.5}{4.6}
& \triple{12.3}{24.7}{10.0}
& \triple{38.1}{17.2}{9.1}
& \triple{27.24}{11.96}{7.7}
\\

& Center-PH
& \triple{\uline{0.97}}{5.9}{\uline{1.1}}
& \triple{\uline{0.88}}{4.8}{\textbf{1.5}}
& \triple{\uline{1.67}}{\uline{13.5}}{\textbf{1.2}}
& \triple{\uline{2.28}}{25.7}{5.2}
& \triple{\textbf{1.11}}{\uline{3.24}}{\uline{5.6}}
\\

& Vanilla
& \triple{4.65}{\uline{4.4}}{3.9}
& \triple{3.24}{14.2}{5.4}
& \triple{9.12}{27.4}{31.6}
& \triple{6.17}{19.7}{38.6}
& \triple{15.36}{11.63}{28.7}
\\

& \textbf{\ourMethod (ours)}
& \triple{\textbf{0.60}}{\textbf{0.7}}{\textbf{0.3}}
& \triple{\textbf{0.76}}{\textbf{1.9}}{\uline{1.9}}
& \triple{\textbf{1.14}}{\textbf{6.4}}{\uline{2.0}}
& \triple{\textbf{0.78}}{\textbf{1.9}}{\textbf{0.9}}
& \triple{\uline{3.49}}{\textbf{2.24}}{\uline{5.6}}
\\

\midrule

\multirow{5}{*}{VGGT~\cite{wang2025vggt}}
& LoRA
& \triple{1.40}{\uline{6.4}}{\uline{1.4}}
& \triple{\uline{0.60}}{25.3}{3.2}
& \triple{13.0}{24.7}{7.9}
& \triple{8.95}{17.2}{\uline{4.4}}
& \triple{6.48}{\uline{3.81}}{\textbf{6.1}}
\\

& CalTok
& \triple{2.45}{15.6}{1.9}
& \triple{\textbf{0.52}}{28.5}{4.0}
& \triple{23.4}{32.2}{9.4}
& \triple{16.7}{26.3}{6.9}
& \triple{7.83}{36.39}{\uline{7.0}}
\\

& Center-PH
& \triple{\uline{0.97}}{7.4}{2.9}
& \triple{0.82}{\uline{4.3}}{\uline{3.0}}
& \triple{\uline{1.13}}{\uline{10.5}}{\textbf{2.4}}
& \triple{\uline{2.45}}{27.3}{6.1}
& \triple{\uline{6.43}}{10.11}{7.58}
\\

& Vanilla
& \triple{5.98}{6.9}{12.4}
& \triple{2.72}{20.9}{14.5}
& \triple{8.54}{22.1}{88.6}
& \triple{7.21}{\uline{16.6}}{39.4}
& \triple{25.31}{27.02}{33.9}
\\

& \textbf{\ourMethod (ours)}
& \triple{\textbf{0.53}}{\textbf{1.0}}{\textbf{1.2}}
& \triple{0.70}{\textbf{3.9}}{\textbf{2.2}}
& \triple{\textbf{1.12}}{\textbf{7.7}}{\uline{3.6}}
& \triple{\textbf{0.93}}{\textbf{6.0}}{\textbf{3.2}}
& \triple{\textbf{6.19}}{\textbf{3.72}}{\textbf{6.1}}
\\

\bottomrule
\end{tabularx}

\end{table}

}

\inparagraph{Dense depth evaluation.}
For ETH3D and ScanNet++, which provide ground-truth depth, we report AbsRel and \(\delta_{1.25}\) in \cref{tab:depth_regime}~(left).
\ourMethod improves over the vanilla model and lightweight adaptation baselines on both datasets.
Center-PH remains strong on depth because it produces perspective images close to the backbone's pretraining distribution.
However, Center-PH discards the peripheral fisheye content, yielding less accurate pose estimates: On ScanNet++, \ourMethod reduces Center-PH rotation from \(3.27^\circ\) to \(1.11^\circ\) and translation from \(22.77^\circ\) to \(5.78^\circ\).

\begin{table}[t]
\centering
\caption{\textbf{Left: Dense depth on ETH3D and ScanNet++.} We report the mean of per-scene means for \textit{AbsRel} (lower is better) and $\delta_{1.25}$ (higher is better).
\textbf{Right: Lightweight adaptation choices on ETH3D.} These results select the LoRA and CalTok settings used in the main comparison. All experiments use DA3-Small as the baseline.
}
\label{tab:depth_regime}
\scriptsize
\setlength{\tabcolsep}{3pt}

\begin{subtable}[t]{0.50\linewidth}
\centering
\begin{tabularx}{\linewidth}{@{}>{\raggedright\arraybackslash}Xcccc@{}}
\toprule
& \multicolumn{2}{c}{ETH3D} & \multicolumn{2}{c}{ScanNet++} \\
\cmidrule(lr){2-3} \cmidrule(lr){4-5} 
Method & AbsRel $\downarrow$ & $\delta_{1.25} \uparrow$ & AbsRel $\downarrow$ & $\delta_{1.25} \uparrow$ \\
\midrule
Center-PH  & \uline{0.111}          & \uline{0.867}          & \textbf{0.066} & \textbf{0.961} \\
LoRA    & 0.166          & 0.814          & 0.175          & 0.760 \\
CalTok4 & 0.175          & 0.793          & 0.168          & 0.769 \\
Vanilla & 0.178          & 0.751          & 0.282          & 0.601 \\
\textbf{\ourMethod (ours)}    & \textbf{0.107} & \textbf{0.884} & \uline{0.108}          & \uline{0.886} \\
\bottomrule
\end{tabularx}
\end{subtable}
\hfill
\begin{subtable}[t]{0.47\linewidth}
\centering
\begin{tabularx}{\linewidth}{@{}>{\raggedright\arraybackslash}Xccc@{}}
\toprule
Method & $R^{\circ}\downarrow$ & $t^{\circ}\downarrow$ & $d_{\mathrm{reproj}}\downarrow$ \\
\midrule
Center-PH                  & 3.46 & 13.70 & 10.92 \\
CalTok $t{=}4$ (default)          & 2.48 & 13.21 & 11.94 \\
CalTok $t{=}8$          & 2.63 & 12.83 & 15.77 \\
LoRA $r{=}4,\alpha{=}8$   & 2.47 & 12.18 & 10.85 \\
LoRA $r{=}8,\alpha{=}16$ (default)  & 2.18 & 10.74 & 9.02 \\
LoRA $r{=}16,\alpha{=}32$ & 3.01 & 11.99 & 10.68 \\
\textbf{\ourMethod (ours)}& 0.70 & 4.48  & 5.82 \\
\bottomrule
\end{tabularx}
\end{subtable}
\vspace{-0.8em}
\end{table}

\inparagraph{Adaptation regimes.}
\cref{tab:depth_regime}~(right) compares lightweight adaptation choices on ETH3D.
LoRA is the strongest generic PEFT baseline, with \(r=8,\alpha=16\) giving the best overall trade-off, while CalTok is sensitive to the number of learned tokens and is best represented by \(t=4\).
Both baselines improve over vanilla inference in some metrics, but their gains are less consistent than \ourMethod.

\inparagraph{Discussion.}
These results support the PE-Jacobian analysis in \cref{sec:prelim}: Fisheye transfer changes the position-to-ray mapping, so camera adaptation is better targeted through positional representations than generic feature transformations.
Although the adapter is fitted on a specific sequence and can absorb some sequence-specific signal, its trainable capacity is restricted to PE/RoPE residuals, biasing adaptation toward camera geometry.

\subsection{Design choices and runtime}
\label{sec:inference_cost}
\cref{tab:component_ablation_sub} isolates the contribution of \ourMethod's components.
The largest gain comes from the learned PE residual. \emph{Radial PE only} already reduces reprojection error substantially, and \emph{Radial + angular PE} further improves the pose--depth balance, indicating that the remaining mismatch is not purely radial.
The \emph{Naive remap of PE} baseline tests whether one can simply reuse the pretrained PE table at undistorted coordinates. It improves rotation, but gives poor translation and depth, showing that direct geometric re-indexing of the pretrained table does not produce consistent fisheye geometry.
The parameter-free components address other places where pinhole assumptions enter the model. \emph{Patch undistortion (no learnable PE)} helps, confirming that tokenization matters, but it remains far below the learned positional adapter. Prediction-grid coordinate correction, border-token handling, and radial RoPE correction provide smaller changes; the \emph{Full model} gives the lowest reprojection error, although some ablations achieve slightly lower pose error.

The extended ablation in \cref{sec:component_ablation_extended} shows the same pattern on ETH3D and FIORD: Learned PE/RoPE residuals drive most of the improvement, while parameter-free corrections provide smaller, dataset-dependent changes.

\cref{tab:runtime_sub} reports inference cost after adaptation.
\ourMethod adds only small positional lookup tables: 20 radial and 8 angular bins for PE, plus 20 radial RoPE bins when applicable.
These tables are evaluated inside the original forward pass, giving negligible overhead over DA3-Small at inference time.
By contrast, \textsc{Multi-PH} requires five virtual pinhole views and costs about \(4\times\) more per frame; LoRA adds around \(10\%\) latency and uses \(\sim\!14\times\) more trainable parameters, while CalTok has 18.4K trainable parameters with modest overhead.
\cref{sec:compute_resources} reports adaptation and evaluation compute.

\begin{table}[t]
\centering
\caption{\textbf{Component ablation and inference cost.}
(\subref{tab:component_ablation_sub}) Component ablation on KITTI-360 (drive 0000, cam02, $185^\circ$ fisheye), trained on 30 frames and evaluated on the first 500 frames.
(\subref{tab:runtime_sub}) Inference cost on DA3-Small, measured per fisheye frame at $504{\times}504$ on one NVIDIA RTX A4000 and averaged over 1000 frames. Trainable parameter counts account for the DA3-Small backbone width, with token size $C=384$.}
\label{tab:ablation_runtime}
\scriptsize
\setlength{\tabcolsep}{3pt}

\begin{subtable}[t]{0.53\linewidth}
\centering
\caption{Component ablation on KITTI-360.}
\label{tab:component_ablation_sub}
\begin{tabularx}{\linewidth}{@{}>{\raggedright\arraybackslash}Xccc@{}}
\toprule
Configuration & $R^{\circ}\downarrow$ & $t^{\circ}\downarrow$ & $d_{\text{reproj}}\downarrow$ \\
\midrule
Patch undistortion (no learnable PE) & 1.397 & 6.66 & 8.96 \\
Naive remap of PE                    & 0.810 & 12.93 & 11.53 \\
Radial PE only                       & 1.154 & 5.48 & 3.70 \\
Radial + angular PE                  & 1.038 & 4.21 & 3.39 \\
\ourMethod w/o border token                & 1.061 & 4.45 & 3.17 \\
\ourMethod w/o DPT pos fix                 & 1.094 & 4.78 & 3.64 \\
\ourMethod w/o RoPE adapter                & 0.942 & 5.56 & 3.27 \\
\textbf{\ourMethod (ours)}                          & 1.183 & 4.81 & 3.03 \\
\bottomrule
\end{tabularx}
\end{subtable}
\hfill
\begin{subtable}[t]{0.44\linewidth}
\centering
\caption{Inference cost on DA3-Small}
\label{tab:runtime_sub}
\begin{tabularx}{\linewidth}{@{}>{\raggedright\arraybackslash}Xccc@{}}
\toprule
Method & ms/frame & Overhead & Params \\
\midrule
Vanilla DA3                    & $\sim$100 & baseline      & 0 \\
Center-PH (single $110^\circ$) & $\sim$105 & $+5\%$        & 0 \\
Multi-PH (5 views)             & $\sim$400 & $+300\%$      & 0 \\
LoRA ($r{=}8$)                 & $\sim$110 & $+10\%$       & 147.5K \\
CalTok ($t{=}4$)               & $\sim$105 & $+5\%$        & 18.4K \\
\midrule
\textbf{\ourMethod (ours)}              & $\sim$100 & $\approx 0\%$ & 10.8K \\
\bottomrule
\end{tabularx}
\end{subtable}

\end{table}

\section{Conclusion and Limitations}

\ourMethod adapts 3D foundation models to fisheye images while preserving the priors acquired from pinhole pretraining.
The method corrects the components linking token locations in the image to geometry: Primarily, absolute positional embeddings, with additional RoPE, prediction-grid, and patch-tokenization corrections.
Across fisheye datasets and frozen backbones, \ourMethod improves pose accuracy while remaining parameter- and runtime-efficient.

\inparagraph{Limitations.}
\label{sec:limitations}
\textit{(i)} \ourMethod computes a camera-specific correction; a different fisheye camera or distortion profile requires new adaptation. 
\textit{(ii)} \ourMethod assumes a principal point and mostly radial distortion; it does not explicitly cover strong tangential or non-radial optics.
\textit{(iii)} \ourMethod requires camera parameters, although \cref{sec:anycalib} shows that predicted calibration from off-the-shelf networks such as AnyCalib~\citep{tiradogarin2025anycalib} remains accurate.
\textit{(iv)} Our work focuses on fisheye images. Panoramic or equirectangular inputs remain for future work.
\textit{(v)} The training set for \ourMethod needs sufficient inter-frame displacement. 
With small or degenerate motion, the self-supervised constraints become weak because large depth or translation-direction errors can induce only small reprojection errors.

Broader impact and asset-license details are provided in \cref{sec:broader_impact,sec:licenses}.

\bibliographystyle{unsrtnat}
\bibliography{main_new}

\begin{thebibliography}{53}
\providecommand{\natexlab}[1]{#1}
\providecommand{\url}[1]{\texttt{#1}}
\expandafter\ifx\csname urlstyle\endcsname\relax
  \providecommand{\doi}[1]{doi: #1}\else
  \providecommand{\doi}{doi: \begingroup \urlstyle{rm}\Url}\fi

\bibitem[Wang et~al.(2024)Wang, Leroy, Cabon, Chidlovskii, and
  Revaud]{wang2024dust3r}
Shuzhe Wang, Vincent Leroy, Yohann Cabon, Boris Chidlovskii, and Jerome Revaud.
\newblock {DUSt3R}: Geometric {3D} vision made easy.
\newblock In \emph{CVPR}, pages 20697--20709, 2024.

\bibitem[Duisterhof et~al.(2025)Duisterhof, Zust, Weinzaepfel, Leroy, Cabon,
  and Revaud]{duisterhof2025mastrsfm}
Bardienus~Pieter Duisterhof, Lojze Zust, Philippe Weinzaepfel, Vincent Leroy,
  Yohann Cabon, and Jerome Revaud.
\newblock {MASt3R-SfM}: A fully integrated solution for unconstrained
  structure-from-motion.
\newblock In \emph{3DV}, 2025.

\bibitem[Wang et~al.(2025)Wang, Chen, Karaev, Vedaldi, Rupprecht, and
  Novotny]{wang2025vggt}
Jianyuan Wang, Minghao Chen, Nikita Karaev, Andrea Vedaldi, Christian
  Rupprecht, and David Novotny.
\newblock {VGGT}: Visual geometry grounded transformer.
\newblock In \emph{CVPR}, pages 5294--5306, 2025.

\bibitem[Wang et~al.(2026)Wang, Zhou, Zhu, Chang, Zhou, Li, Chen, Pang, Shen,
  and He]{wang2026pi3permutationequivariantvisualgeometry}
Yifan Wang, Jianjun Zhou, Haoyi Zhu, Wenzheng Chang, Yang Zhou, Zizun Li, Junyi
  Chen, Jiangmiao Pang, Chunhua Shen, and Tong He.
\newblock {$\pi^3$}: Permutation-equivariant visual geometry learning.
\newblock In \emph{ICLR}, 2026.

\bibitem[Lin et~al.(2026)Lin, Chen, Liew, Chen, Li, Zhao, Peng, Guo, Zhou, Shi,
  Feng, and Kang]{lin2025depth}
Haotong Lin, Sili Chen, Junhao Liew, Donny~Y. Chen, Zhenyu Li, Yang Zhao, Sida
  Peng, Hengkai Guo, Xiaowei Zhou, Guang Shi, Jiashi Feng, and Bingyi Kang.
\newblock {Depth Anything 3}: Recovering the visual space from any views.
\newblock In \emph{ICLR}, 2026.

\bibitem[Wang et~al.(2018)Wang, Cai, Li, Liu, Guo, Li, and
  Cheng]{wang2018cubemapslam}
Yahui Wang, Shaojun Cai, Shi-Jie Li, Yun Liu, Yangyan Guo, Tao Li, and
  Ming-Ming Cheng.
\newblock {CubemapSLAM}: A piecewise-pinhole monocular fisheye {SLAM} system.
\newblock In \emph{ACCV}, pages 34--49, 2018.

\bibitem[Xu et~al.(2024)Xu, Yin, Gong, Jiang, and Liu]{xu2024sdge}
Jialei Xu, Wei Yin, Dong Gong, Junjun Jiang, and Xianming Liu.
\newblock {SDGE}: Stereo guided depth estimation for 360 camera sets.
\newblock In \emph{IROS}, pages 11179--11186, 2024.

\bibitem[Hu et~al.(2022)Hu, Shen, Wallis, Allen-Zhu, Li, Wang, Wang, Chen,
  et~al.]{hu2022lora}
Edward~J. Hu, Yelong Shen, Phillip Wallis, Zeyuan Allen-Zhu, Yuanzhi Li, Shean
  Wang, Liang Wang, Weizhu Chen, et~al.
\newblock {LoRA}: Low-rank adaptation of large language models.
\newblock In \emph{ICLR}, 2022.

\bibitem[Houlsby et~al.(2019)Houlsby, Giurgiu, Jastrzebski, Morrone,
  De~Laroussilhe, Gesmundo, Attariyan, and Gelly]{houlsby2019parameter}
Neil Houlsby, Andrei Giurgiu, Stanislaw Jastrzebski, Bruna Morrone, Quentin
  De~Laroussilhe, Andrea Gesmundo, Mona Attariyan, and Sylvain Gelly.
\newblock Parameter-efficient transfer learning for {NLP}.
\newblock In \emph{ICML}, pages 2790--2799, 2019.

\bibitem[Lian et~al.(2022)Lian, Zhou, Feng, and Wang]{lian2022scaling}
Dongze Lian, Daquan Zhou, Jiashi Feng, and Xinchao Wang.
\newblock Scaling and shifting your features: A new baseline for efficient
  model tuning.
\newblock In \emph{NeurIPS}, 2022.

\bibitem[Gangopadhyay et~al.(2025)Gangopadhyay, Kim, Chen, Rim, Park, and
  Wong]{gangopadhyay2025extending}
Suchisrit Gangopadhyay, Jung-Hee Kim, Xien Chen, Patrick Rim, Hyoungseob Park,
  and Alex Wong.
\newblock Extending foundational monocular depth estimators to fisheye cameras
  with calibration tokens.
\newblock In \emph{ICCV}, pages 5198--5209, 2025.

\bibitem[Sch\"{o}nberger and Frahm(2016)]{colmap}
Johannes~Lutz Sch\"{o}nberger and Jan-Michael Frahm.
\newblock Structure-from-motion revisited.
\newblock In \emph{CVPR}, pages 4104--4113, 2016.

\bibitem[Sarlin et~al.(2021)Sarlin, Unagar, Larsson, Germain, Toft, Larsson,
  Pollefeys, Lepetit, Hammarstrand, Kahl, et~al.]{sarlin2021back}
Paul-Edouard Sarlin, Ajaykumar Unagar, Mans Larsson, Hugo Germain, Carl Toft,
  Viktor Larsson, Marc Pollefeys, Vincent Lepetit, Lars Hammarstrand, Fredrik
  Kahl, et~al.
\newblock Back to the feature: Learning robust camera localization from pixels
  to pose.
\newblock In \emph{CVPR}, pages 3247--3257, 2021.

\bibitem[Solonets et~al.(2024)Solonets, Sinitsyn, Von~Stumberg, Araslanov, and
  Cremers]{solonets2024analytical}
Sergei Solonets, Daniil Sinitsyn, Lukas Von~Stumberg, Nikita Araslanov, and
  Daniel Cremers.
\newblock An analytical solution to gauss-newton loss for direct image
  alignment.
\newblock In \emph{ICLR}, 2024.

\bibitem[McCraith et~al.(2020)McCraith, Neumann, Zisserman, and
  Vedaldi]{mccraith2020monocular}
Robert McCraith, Lukas Neumann, Andrew Zisserman, and Andrea Vedaldi.
\newblock Monocular depth estimation with self-supervised instance adaptation.
\newblock \emph{arXiv:2004.05821 [cs.CV]}, 2020.

\bibitem[Xie et~al.(2023)Xie, Wang, and Liu]{xie2023omnividar}
Sheng Xie, Daochuan Wang, and Yun-Hui Liu.
\newblock {OmniViDAR}: Omnidirectional depth estimation from multi-fisheye
  images.
\newblock In \emph{CVPR}, pages 21529--21538, 2023.

\bibitem[Deng et~al.(2025)Deng, Wang, Meng, Hou, Chang, and
  Chen]{deng2025omnistereo}
Jiaxi Deng, Yushen Wang, Haitao Meng, Zuoxun Hou, Yi~Chang, and Gang Chen.
\newblock {OmniStereo}: Real-time omnidirectional depth estimation with
  multi-view fisheye cameras.
\newblock In \emph{CVPR}, pages 1003--1012, 2025.

\bibitem[Guo et~al.(2025)Guo, Garg, Miangoleh, Huang, and Ren]{guo2025depth}
Yuliang Guo, Sparsh Garg, S.~Mahdi~H. Miangoleh, Xinyu Huang, and Liu Ren.
\newblock Depth any camera: Zero-shot metric depth estimation from any camera.
\newblock In \emph{CVPR}, pages 26996--27006, 2025.

\bibitem[Zhao et~al.(2025)Zhao, Liu, Qi, Ma, Liu, and Ma]{zhao2025fisheyedepth}
Guoyang Zhao, Yuxuan Liu, Weiqing Qi, Fulong Ma, Ming Liu, and Jun Ma.
\newblock {FisheyeDepth}: A real-scale self-supervised depth estimation model
  for fisheye camera.
\newblock In \emph{ICRA}, pages 3780--3787, 2025.

\bibitem[Piccinelli et~al.(2024)Piccinelli, Yang, Sakaridis, Segu, Li, Gool,
  and Yu]{piccinelli2024unidepthuniversalmonocularmetric}
Luigi Piccinelli, Yung-Hsu Yang, Christos Sakaridis, Mattia Segu, Siyuan Li,
  Luc~Van Gool, and Fisher Yu.
\newblock {UniDepth}: Universal monocular metric depth estimation.
\newblock In \emph{CVPR}, pages 10106--10116, 2024.

\bibitem[Piccinelli et~al.(2025)Piccinelli, Sakaridis, Segu, Yang, Li,
  Abbeloos, and Van~Gool]{piccinelli2025unik3d}
Luigi Piccinelli, Christos Sakaridis, Mattia Segu, Yung-Hsu Yang, Siyuan Li,
  Wim Abbeloos, and Luc Van~Gool.
\newblock {UniK3D}: Universal camera monocular {3D} estimation.
\newblock In \emph{CVPR}, pages 1028--1039, 2025.

\bibitem[Hu et~al.(2024)Hu, Yin, Zhang, Cai, Long, Chen, Wang, Yu, Shen, and
  Shen]{hu2024metric3dv2}
Mu~Hu, Wei Yin, Chi Zhang, Zhipeng Cai, Xiaoxiao Long, Hao Chen, Kaixuan Wang,
  Gang Yu, Chunhua Shen, and Shaojie Shen.
\newblock {Metric3D v2}: A versatile monocular geometric foundation model for
  zero-shot metric depth and surface normal estimation.
\newblock \emph{IEEE Transactions on Pattern Analysis and Machine
  Intelligence}, 46\penalty0 (12):\penalty0 10579--10596, 2024.

\bibitem[Tirado-Gar{\'\i}n and Civera(2025)]{tiradogarin2025anycalib}
Javier Tirado-Gar{\'\i}n and Javier Civera.
\newblock {AnyCalib}: On-manifold learning for model-agnostic single-view
  camera calibration.
\newblock In \emph{ICCV}, pages 8044--8055, 2025.

\bibitem[Sinitsyn et~al.(2025)Sinitsyn, H{\"a}renstam-Nielsen, and
  Cremers]{prada2025}
Daniil Sinitsyn, Linus H{\"a}renstam-Nielsen, and Daniel Cremers.
\newblock {PRaDA}: Projective radial distortion averaging.
\newblock In \emph{CVPR}, pages 21902--21912, 2025.

\bibitem[Athwale et~al.(2023)Athwale, Afrasiyabi, Lag{\"u}e, Shili, Ahmad, and
  Lalonde]{athwale2023darswin}
Akshaya Athwale, Arman Afrasiyabi, Justin Lag{\"u}e, Ichrak Shili, Ola Ahmad,
  and Jean-Fran{\c{c}}ois Lalonde.
\newblock {DarSwin}: Distortion-aware radial swin transformer.
\newblock In \emph{ICCV}, pages 5929--5938, 2023.

\bibitem[Athwale et~al.(2025)Athwale, Shili, Bergeron, Ahmad, and
  Lalonde]{athwale2025darswin}
Akshaya Athwale, Ichrak Shili, {\'E}mile Bergeron, Ola Ahmad, and
  Jean-Fran{\c{c}}ois Lalonde.
\newblock {DarSwin-UNet}: Distortion-aware architecture.
\newblock In \emph{WACV}, pages 8670--8680, 2025.

\bibitem[Yang et~al.(2023)Yang, Tang, Gao, Yang, and Fu]{yang2023sector}
Dianyi Yang, Jiadong Tang, Yu~Gao, Yi~Yang, and Mengyin Fu.
\newblock Sector patch embedding: An embedding module conforming to the
  distortion pattern of fisheye image.
\newblock \emph{arXiv:2303.14645 [cs.CV]}, 2023.

\bibitem[Li et~al.(2025)Li, Yi, Liu, Gao, Ma, and Kanazawa]{li2025cameras}
Ruilong Li, Brent Yi, Junchen Liu, Hang Gao, Yi~Ma, and Angjoo Kanazawa.
\newblock Cameras as relative positional encoding.
\newblock In \emph{NeurIPS}, 2025.

\bibitem[Duan et~al.(2026)Duan, Hong, Zhao, Turner, Wong, and
  Zhou]{duan2026fisheye3r}
Ruxiao Duan, Erin Hong, Dongxu Zhao, Eric Turner, Alex Wong, and Yunwen Zhou.
\newblock {Fisheye3R}: Adapting unified {3D} feed-forward foundation models to
  fisheye lenses.
\newblock \emph{arXiv:2603.28896 [cs.CV]}, 2026.

\bibitem[Ahuja et~al.(2026)Ahuja, Jain, Sudhakar, Narayanan, Likhar, Kumar, and
  Yogamani]{ahuja2026fishropeprojectiverotaryposition}
Rahul Ahuja, Mudit Jain, Bala Murali Manoghar~Sai Sudhakar, Venkatraman
  Narayanan, Pratik Likhar, Varun~Ravi Kumar, and Senthil Yogamani.
\newblock {FishRoPE}: Projective rotary position embeddings for omnidirectional
  visual perception.
\newblock \emph{arXiv:2604.10391 [cs.CV]}, 2026.

\bibitem[Wang and Liu(2024)]{wang2024depthanywhere}
Ning-Hsu Wang and Yu-Lun Liu.
\newblock {Depth Anywhere}: Enhancing 360 monocular depth estimation via
  perspective distillation and unlabeled data augmentation.
\newblock In \emph{NeurIPS}, 2024.

\bibitem[Yuan et~al.(2026)Yuan, Jiang, Soh, and Zhao]{yuan2026vggt360}
Jiayi Yuan, Haobo Jiang, De~Wen Soh, and Na~Zhao.
\newblock {VGGT-360}: Geometry-consistent zero-shot panoramic depth estimation.
\newblock \emph{arXiv:2603.18943 [cs.CV]}, 2026.

\bibitem[Jung et~al.(2025)Jung, Choi, Lee, and Manocha]{jung2025rpg360}
Dongki Jung, Jaehoon Choi, Yonghan Lee, and Dinesh Manocha.
\newblock {RPG360}: Robust 360 depth estimation with perspective foundation
  models and graph optimization.
\newblock In \emph{NeurIPS}, 2025.

\bibitem[Sun et~al.(2020)Sun, Wang, Liu, Miller, Efros, and Hardt]{sun2020test}
Yu~Sun, Xiaolong Wang, Zhuang Liu, John Miller, Alexei Efros, and Moritz Hardt.
\newblock Test-time training with self-supervision for generalization under
  distribution shifts.
\newblock In \emph{ICML}, pages 9229--9248, 2020.

\bibitem[Zhang et~al.(2026)Zhang, Bi, Hong, Zhang, Luan, Yang, Sunkavalli,
  Freeman, and Tan]{zhang2025testtimetrainingright}
Tianyuan Zhang, Sai Bi, Yicong Hong, Kai Zhang, Fujun Luan, Songlin Yang,
  Kalyan Sunkavalli, William~T. Freeman, and Hao Tan.
\newblock Test-time training done right.
\newblock In \emph{ICLR}, 2026.

\bibitem[Chen et~al.(2026)Chen, Chen, Xiu, Geiger, and Chen]{chen2025ttt3r}
Xingyu Chen, Yue Chen, Yuliang Xiu, Andreas Geiger, and Anpei Chen.
\newblock {TTT3R}: {3D} reconstruction as test-time training.
\newblock In \emph{ICLR}, 2026.

\bibitem[Xie et~al.(2026)Xie, Yang, Jin, Cai, Yin, Ren, Zhang, Hua, Peng, Guo,
  and Zhou]{xie2026scal3rscalabletesttimetraining}
Tao Xie, Peishan Yang, Yudong Jin, Yingfeng Cai, Wei Yin, Weiqiang Ren, Qian
  Zhang, Wei Hua, Sida Peng, Xiaoyang Guo, and Xiaowei Zhou.
\newblock {Scal3R}: Scalable test-time training for large-scale {3D}
  reconstruction.
\newblock \emph{arXiv:2604.08542 [cs.CV]}, 2026.

\bibitem[Kannala and Brandt(2006)]{kannala2006generic}
Juho Kannala and Sami~S. Brandt.
\newblock A generic camera model and calibration method for conventional,
  wide-angle, and fish-eye lenses.
\newblock \emph{IEEE Transactions on Pattern Analysis and Machine
  Intelligence}, 28\penalty0 (8):\penalty0 1335--1340, 2006.

\bibitem[Khomutenko et~al.(2015)Khomutenko, Garcia, and
  Martinet]{khomutenko2015enhanced}
Bogdan Khomutenko, Ga{\"e}tan Garcia, and Philippe Martinet.
\newblock An enhanced unified camera model.
\newblock \emph{IEEE Robotics and Automation Letters}, 1\penalty0 (1):\penalty0
  137--144, 2015.

\bibitem[Oquab et~al.(2024)Oquab, Darcet, Moutakanni, Vo, Szafraniec, Khalidov,
  Fernandez, Haziza, Massa, El-Nouby, Assran, Ballas, Galuba, Howes, Huang, Li,
  Misra, Rabbat, Sharma, Synnaeve, Xu, Jegou, Mairal, Labatut, Joulin, and
  Bojanowski]{oquab2024dinov2learningrobustvisual}
Maxime Oquab, Timothée Darcet, Théo Moutakanni, Huy Vo, Marc Szafraniec,
  Vasil Khalidov, Pierre Fernandez, Daniel Haziza, Francisco Massa, Alaaeldin
  El-Nouby, Mahmoud Assran, Nicolas Ballas, Wojciech Galuba, Russell Howes,
  Po-Yao Huang, Shang-Wen Li, Ishan Misra, Michael Rabbat, Vasu Sharma, Gabriel
  Synnaeve, Hu~Xu, Hervé Jegou, Julien Mairal, Patrick Labatut, Armand Joulin,
  and Piotr Bojanowski.
\newblock {DINOv2}: Learning robust visual features without supervision.
\newblock \emph{TMLR}, 2024.

\bibitem[Heo et~al.(2024)Heo, Park, Han, and Yun]{heo2024rotary}
Byeongho Heo, Song Park, Dongyoon Han, and Sangdoo Yun.
\newblock Rotary position embedding for vision transformer.
\newblock In \emph{ECCV}, pages 289--305, 2024.

\bibitem[Ranftl et~al.(2021)Ranftl, Bochkovskiy, and
  Koltun]{ranftl2021visiontransformersdenseprediction}
René Ranftl, Alexey Bochkovskiy, and Vladlen Koltun.
\newblock Vision transformers for dense prediction.
\newblock In \emph{ICCV}, pages 12179--12188, 2021.

\bibitem[Qin and Li(2012)]{qin2012tracking}
Xuebin Qin and Shigang Li.
\newblock Tracking feature points of fisheye full-view image by normalized
  image patch.
\newblock \emph{IEEJ Transactions on Electronics, Information and Systems},
  132\penalty0 (9):\penalty0 1516--1523, 2012.

\bibitem[Zhang et~al.(2025)Zhang, Keetha, Lyu, Jhamb, Chen, Qiu, Karhade, Jha,
  Hu, Ramanan, Scherer, and Wang]{zhang2025ufm}
Yuchen Zhang, Nikhil Keetha, Chenwei Lyu, Bhuvan Jhamb, Yutian Chen, Yuheng
  Qiu, Jay Karhade, Shreyas Jha, Yaoyu Hu, Deva Ramanan, Sebastian Scherer, and
  Wenshan Wang.
\newblock {UFM}: A simple path towards unified dense correspondence with flow.
\newblock In \emph{NeurIPS}, 2025.

\bibitem[Barath et~al.(2020)Barath, Noskova, Ivashechkin, and
  Matas]{barath2020magsac++}
Daniel Barath, Jana Noskova, Maksym Ivashechkin, and Jiri Matas.
\newblock {MAGSAC++}: A fast, reliable, and accurate robust estimator.
\newblock In \emph{CVPR}, pages 1304--1312, 2020.

\bibitem[Godard et~al.(2017)Godard, Mac~Aodha, and
  Brostow]{godard2017unsupervised}
Cl{\'e}ment Godard, Oisin Mac~Aodha, and Gabriel~J. Brostow.
\newblock Unsupervised monocular depth estimation with left-right consistency.
\newblock In \emph{CVPR}, pages 270--279, 2017.

\bibitem[Wimbauer et~al.(2025)Wimbauer, Chen, Muhle, Rupprecht, and
  Cremers]{wimbauer2025anycamlearningrecovercamera}
Felix Wimbauer, Weirong Chen, Dominik Muhle, Christian Rupprecht, and Daniel
  Cremers.
\newblock {AnyCam}: Learning to recover camera poses and intrinsics from casual
  videos.
\newblock In \emph{CVPR}, pages 16717--16727, 2025.

\bibitem[Liao et~al.(2023)Liao, Xie, and
  Geiger]{liao2022kitti360noveldatasetbenchmarks}
Yiyi Liao, Jun Xie, and Andreas Geiger.
\newblock {KITTI-360}: A novel dataset and benchmarks for urban scene
  understanding in {2D} and {3D}.
\newblock \emph{IEEE Transactions on Pattern Analysis and Machine
  Intelligence}, 45\penalty0 (3):\penalty0 3292--3310, 2023.

\bibitem[Schubert et~al.(2018)Schubert, Goll, Demmel, Usenko, Stuckler, and
  Cremers]{Schubert_2018}
David Schubert, Thore Goll, Nikolaus Demmel, Vladyslav Usenko, Jorg Stuckler,
  and Daniel Cremers.
\newblock The {TUM VI} benchmark for evaluating visual-inertial odometry.
\newblock In \emph{IROS}, pages 1680--1687, 2018.

\bibitem[Yeshwanth et~al.(2023)Yeshwanth, Liu, Nie{\ss}ner, and
  Dai]{yeshwanth2023scannethighfidelitydataset3d}
Chandan Yeshwanth, Yueh-Cheng Liu, Matthias Nie{\ss}ner, and Angela Dai.
\newblock {ScanNet++}: A high-fidelity dataset of {3D} indoor scenes.
\newblock In \emph{ICCV}, pages 12--22, 2023.

\bibitem[Schops et~al.(2017)Schops, Schonberger, Galliani, Sattler, Schindler,
  Pollefeys, and Geiger]{schops2017multi}
Thomas Schops, Johannes~L. Schonberger, Silvano Galliani, Torsten Sattler,
  Konrad Schindler, Marc Pollefeys, and Andreas Geiger.
\newblock A multi-view stereo benchmark with high-resolution images and
  multi-camera videos.
\newblock In \emph{CVPR}, pages 3260--3269, 2017.

\bibitem[Gunes et~al.(2025)Gunes, Turkulainen, Ren, Solin, Kannala, and
  Rahtu]{gunes2025fiordfisheyeindooroutdoordataset}
Ulas Gunes, Matias Turkulainen, Xuqian Ren, Arno Solin, Juho Kannala, and Esa
  Rahtu.
\newblock {FIORD}: A fisheye indoor-outdoor dataset with {LiDAR} ground truth
  for {3D} scene reconstruction and benchmarking.
\newblock In \emph{SCIA}, pages 3--17, 2025.

\bibitem[Eigen et~al.(2014)Eigen, Puhrsch, and Fergus]{eigen2014depth}
David Eigen, Christian Puhrsch, and Rob Fergus.
\newblock Depth map prediction from a single image using a multi-scale deep
  network.
\newblock In \emph{NeurIPS}, 2014.

\end{thebibliography}

\appendix
\clearpage
\pagenumbering{roman}

\section{Evaluation metrics}
\label{sec:err_metrics}

For each frame pair \((i,j)\), let \((\hat{R}_{ij},\hat{t}_{ij})\) be the predicted relative pose and \((R^\star_{ij},t^\star_{ij})\) the ground-truth pose. We report angular rotation and translation-direction errors in degrees:
\begin{equation}
R_{\mathrm{err}}(i,j)=
\arccos\!\left(
\frac{\mathrm{tr}\!\left((R^\star_{ij})^\top \hat{R}_{ij}\right)-1}{2}
\right),
\qquad
t_{\mathrm{err}}(i,j)=
\arccos\!\left(
\frac{\langle t^\star_{ij},\hat{t}_{ij}\rangle}
{\|t^\star_{ij}\|\,\|\hat{t}_{ij}\|}
\right).
\end{equation}

Ground-truth depth is not available for all fisheye datasets and is often sparse or sensor-projected.
Therefore, we use a reprojection-based depth metric for the main comparison. Let
\begin{equation}
X_i(u)=D_i(u)\,\kappa^{-1}(u)
\end{equation}
be the 3D point induced by predicted depth at pixel \(u\). For each image pair, we resolve the global scale ambiguity with a single scalar \(s\), and define
\begin{equation}
d_{\mathrm{reproj}}(i,j)=
\min_{s>0}\;
\frac{1}{|\Omega|}
\sum_{u\in\Omega}
\left\|
\kappa\!\left(R^\star_{ij}\bigl(s\,X_i(u)\bigr)+t^\star_{ij}\right)-u_{ij}(u)
\right\|_2 .
\end{equation}
$d_{\mathrm{reproj}}$ uses ground-truth pose, so it evaluates depth independently of predicted pose. In practice, \(s\) is estimated per pair in closed form by robustly minimizing the distance between high-confidence matched 3D points.

When reliable ground-truth depth \(D_i^\star\) is available, we additionally report scale-aligned depth error following~\citet{eigen2014depth}. With a scalar \(s_i^\star\) fitted to the ground-truth depth, we define
\begin{equation}
d_{AbsRel}(i)=
\frac{1}{|\Omega_i|}
\sum_{u\in\Omega_i}
\left\|
s_i^\star D_i(u)-D_i^\star(u)
\right\|_2 .
\end{equation}

We also report threshold accuracy \(\delta_{1.25}\), the fraction of pixels whose predicted depth is within a factor \(1.25\) of ground truth:
\begin{equation}
\delta_{1.25}(i)=
\frac{1}{|\Omega_i|}
\sum_{u\in\Omega_i}
\mathbf{1}\!\left(
\max\!\left(
\frac{s_i^\star D_i(u)}{D_i^\star(u)},
\frac{D_i^\star(u)}{s_i^\star D_i(u)}
\right)<1.25
\right).
\end{equation}

\section{Predicted calibration}
\label{sec:anycalib}

{
\newcolumntype{M}{>{\centering\arraybackslash}X}
\newcommand{\triple}[3]{%
  \makebox[1.85em][r]{#1}%
  \hspace{0.25em}%
  \makebox[1.85em][r]{#2}%
  \hspace{0.25em}%
  \makebox[1.85em][r]{#3}%
}

\begin{table}[b]
\centering
\caption{\textbf{Ground-truth vs.\ AnyCalib calibration with DA3-Small.}
We evaluate one representative sequence per dataset. 
For each table entry we report the triple comprising $R^{\circ}$, $t^{\circ}$, and $d_{\mathrm{reproj}}$ (lower is better). \ourMethod outperforms both LoRA and CalTok \emdash often by a significant margin \emdash also with approximate camera calibration estimated by AnyCalib.}
\label{tab:anycalib_vs_gt}
\scriptsize
\setlength{\tabcolsep}{2.5pt}

\begin{tabularx}{\linewidth}{@{}lMMMMM@{}}
\toprule
Method
& \makecell{ETH3D\\terrains}
& \makecell{KITTI\\0009}
& \makecell{TUM-VI\\room6}
& \makecell{ScanNet++\\3f15}
& \makecell{FIORD\\Kitchen} \\
\midrule

\ourMethod (ours) (GT)
& \triple{0.48}{0.9}{1.7}
& \triple{0.69}{2.4}{2.9}
& \triple{1.94}{11.2}{3.6}
& \triple{0.40}{2.2}{1.7}
& \triple{3.1}{2.5}{5.5} \\

\ourMethod (ours) (AnyCalib)
& \triple{1.02}{2.3}{2.6}
& \triple{1.14}{2.5}{3.1}
& \triple{1.16}{7.0}{4.4}
& \triple{0.75}{4.6}{3.1}
& \triple{4.9}{3.0}{5.2} \\

\midrule

LoRA (GT)
& \triple{3.62}{3.4}{4.0}
& \triple{0.61}{2.8}{4.6}
& \triple{2.96}{12.6}{2.8}
& \triple{4.22}{23.0}{2.9}
& \triple{7.7}{14.6}{10.1} \\

LoRA (AnyCalib)
& \triple{2.30}{4.7}{3.2}
& \triple{2.32}{2.8}{4.8}
& \triple{6.60}{20.4}{16.6}
& \triple{3.52}{17.5}{4.9}
& \triple{10.2}{14.2}{10.8} \\

\midrule

CalTok (GT)
& \triple{3.41}{4.8}{4.5}
& \triple{1.09}{3.1}{4.4}
& \triple{3.79}{14.9}{4.6}
& \triple{3.09}{20.0}{4.4}
& \triple{15.8}{15.5}{9.0} \\

CalTok (AnyCalib)
& \triple{2.15}{4.4}{3.4}
& \triple{1.90}{8.5}{5.8}
& \triple{7.21}{42.0}{38.1}
& \triple{4.44}{24.7}{5.5}
& \triple{30.3}{59.9}{10.6} \\

\bottomrule
\end{tabularx}
\end{table}
}

When ground-truth intrinsics are unavailable, we replace them with per-image predictions from AnyCalib~\citep{tiradogarin2025anycalib} on one representative sequence per dataset and re-run adaptation. \cref{tab:anycalib_vs_gt} shows that our method remains useful under approximate camera calibration and outperforms the competitors LoRA and CalTok consistently and significantly across all settings.
The errors vary across sequences; however, \ourMethod exhibits higher robustness compared to the larger failures observed for LoRA and CalTok on TUM-VI, ScanNet++, and FIORD.

\section{Can we simply undistort fisheye patch coordinates?}
\label{sec:patch_undistort}

\begin{figure}[t]
    \centering
        \begin{subfigure}[t]{0.23\linewidth}
        \centering
        \includegraphics[height=0.8\linewidth]{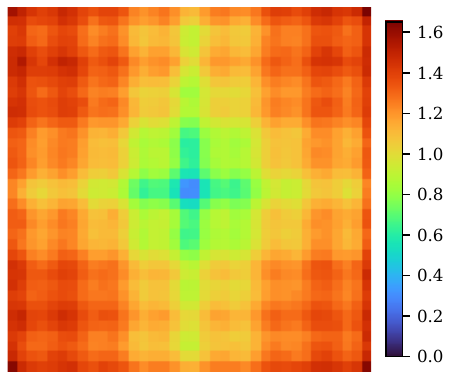}
        \caption{Original PE}
    \end{subfigure}%
    \begin{subfigure}[t]{0.23\linewidth}
        \centering
        \includegraphics[height=0.8\linewidth]{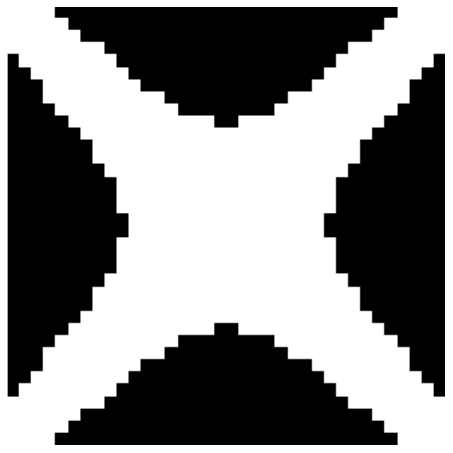}
        \caption{Undistortion pattern}
    \end{subfigure}%
    \begin{subfigure}[t]{0.23\linewidth}
        \centering
        \includegraphics[height=0.8\linewidth]{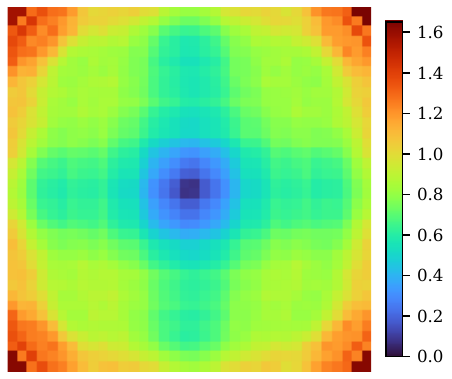}
        \caption{Naive sampling}
    \end{subfigure}
    \caption{\textbf{Naive undistortion-based PE reuse.} (a) Original DA3-small PE table, shown as $\ell_2$ distance to the center embedding. (b) Locations reached by undistorted and rescaled fisheye coordinates. (c) Resulting PE lookup. Many fisheye patches collapse onto a small central region of the table, producing nearly identical embeddings and losing positional resolution.}
    \label{fig:pe_naive_lookup}
\end{figure}

An alternative to adapting the model is to undistort each fisheye patch coordinate and to reuse the original 2D positional embedding table. This fails for wide-angle fisheye images because undistortion samples the pinhole grid highly non-uniformly: Many fisheye patches collapse onto a small central region of the original table, making their embeddings nearly indistinguishable, as \cref{fig:pe_naive_lookup} illustrates. 
The corresponding \emph{Naive remap of PE} row in \cref{tab:component_ablation_sub} confirms that this re-indexing alone gives poor translation and depth.

We therefore learn a lightweight residual correction directly on the positional embeddings, while keeping the pretrained backbone frozen.

\section{Computational resources}
\label{sec:compute_resources}
The experiments were run on NVIDIA RTX A6000 and NVIDIA RTX A4000 GPUs, with independent scenes, sequences, and baselines executed in parallel when possible. A typical full train-and-evaluate run takes about 2--3 hours per ETH3D, ScanNet++, or FIORD scene, and about 3--5 hours per KITTI-360 drive or TUM-VI sequence because of longer sequences. The reported experiments required approximately 180--250 total GPU-hours. Additional preliminary development and failed configurations required extra compute but are not included in this estimate.

\section{ETH3D across backbones}
\label{sec:all_backbones_eth3d}

{
\newcolumntype{M}{>{\centering\arraybackslash}X}
\newcommand{\metriccell}[4]{%
  \makebox[2.5em][r]{#1}%
  \hspace{0.22em}%
  \makebox[2.5em][r]{#2}%
  \hspace{0.22em}%
  \makebox[2.5em][r]{#3}%
  \hspace{0.22em}%
  \makebox[2.5em][r]{#4}%
}
\newcommand{\metrichead}{%
  \makebox[2.5em][c]{$R$}%
  \hspace{0.22em}%
  \makebox[2.5em][c]{$t$}%
  \hspace{0.22em}%
  \makebox[2.5em][c]{AbsRel}%
  \hspace{0.22em}%
  \makebox[2.5em][c]{$\delta$}%
}

\begin{table}[t]
\centering
\caption{\textbf{Mean ETH3D results across backbones.}
For each frozen backbone, entries are ordered as
$R^{\circ}$, $t^{\circ}$, AbsRel, and $\delta_{1.25}$.
Lower is better except for $\delta_{1.25}$.
\textbf{Bold} marks the best method within each backbone and metric.}
\label{tab:eth3d_backbone_means}
\tiny
\setlength{\tabcolsep}{1pt}

\begin{tabularx}{\linewidth}{@{}lMMMMM@{}}
\toprule
Method & DA3-Small & DA3-Base & DA3-Large & \pithree & VGGT \\
\cmidrule(lr){2-2} \cmidrule(lr){3-3} \cmidrule(lr){4-4} \cmidrule(lr){5-5} \cmidrule(lr){6-6}
& \metrichead & \metrichead & \metrichead & \metrichead & \metrichead \\
\midrule

Baseline
& \metriccell{8.59}{15.16}{0.178}{0.751}
& \metriccell{8.27}{12.24}{0.147}{0.794}
& \metriccell{6.36}{13.94}{0.135}{0.828}
& \metriccell{2.66}{11.30}{0.250}{0.642}
& \metriccell{3.19}{11.52}{0.285}{0.557} \\

Center-PH
& \metriccell{3.46}{13.70}{0.111}{0.867}
& \metriccell{1.85}{9.36}{\textbf{0.082}}{\textbf{0.911}}
& \metriccell{1.56}{9.32}{\textbf{0.075}}{\textbf{0.941}}
& \metriccell{1.08}{10.46}{\textbf{0.156}}{0.772}
& \metriccell{1.17}{8.98}{0.228}{0.623} \\

\textbf{\ourMethod (ours)}
& \metriccell{\textbf{0.70}}{\textbf{4.48}}{\textbf{0.107}}{\textbf{0.884}}
& \metriccell{\textbf{0.54}}{\textbf{3.26}}{0.089}{0.910}
& \metriccell{\textbf{0.51}}{\textbf{2.96}}{0.083}{0.925}
& \metriccell{\textbf{0.66}}{\textbf{2.48}}{0.175}{\textbf{0.863}}
& \metriccell{\textbf{0.96}}{\textbf{4.82}}{\textbf{0.139}}{\textbf{0.834}} \\

\bottomrule
\end{tabularx}
\end{table}
}

\cref{tab:eth3d_backbone_means} reports mean ETH3D results across scenes for five frozen backbones: DA3-Small, DA3-Base, DA3-Large, \pithree, and VGGT. 
\ourMethod gives the lowest mean rotation and translation error for every backbone. 
Center-PH is competitive on depth for DA3-Base, DA3-Large, and \pithree, but its pose errors remain consistently higher. 
Vanilla inference degrades more strongly under fisheye geometry, especially for pose.

\section{Loss-component ablation}
\label{sec:ablation_loss}

To justify the adaptation objective in \cref{eq:training_loss}, we run an ablation on the ETH3D \texttt{terrains} sequence with different combinations of the loss components. 
\cref{tab:loss_ablation} shows that the full objective is best or tied on this sequence.
Removing the pose loss mainly degrades pose accuracy, while removing either the \(\ell_2\) or TV regularizer generally weakens pose or reprojection performance.

\begin{table}[t]
\centering
\caption{\textbf{Loss, PE/RoPE, and bin-count ablations.}
(\subref{tab:loss_ablation}) Ablation of loss components; we study the effect of pose supervision and the two PE regularizers from \cref{eq:training_loss}.
(\subref{tab:pe_rope_bins}) PE/RoPE split and bin-count sensitivity on ETH3D terrains; each entry reports $R^\circ, t^\circ, d_{\mathrm{reproj}}$, respectively; lower is better.}
\label{tab:loss_pe_ablation}
\scriptsize
\setlength{\tabcolsep}{3pt}

\begin{subtable}[t]{0.49\linewidth}
\centering
\caption{Ablation of loss components.}
\label{tab:loss_ablation}

\newcolumntype{Y}{>{\centering\arraybackslash}X}

\begin{tabularx}{\linewidth}{@{}YYYccc@{}}
\toprule
\multicolumn{3}{c}{Loss components} &
\multicolumn{3}{c}{ETH3D terrains} \\
\cmidrule(lr){1-3}
\cmidrule(l){4-6}
Pose & L2 & TV
& $R^\circ\downarrow$
& $t^\circ\downarrow$
& $d_{\mathrm{reproj}}\downarrow$ \\
\midrule
\checkmark & \checkmark & \checkmark & 0.48 & 0.9 & 1.7 \\
\texttimes & \checkmark & \checkmark & 0.68 & 1.1 & 1.7 \\
\checkmark & \texttimes & \checkmark & 0.81 & 1.0 & 1.8 \\
\checkmark & \checkmark & \texttimes & 0.58 & 1.3 & 1.9 \\
\texttimes & \texttimes & \checkmark & 0.65 & 1.4 & 1.8 \\
\texttimes & \checkmark & \texttimes & 0.61 & 1.5 & 1.7 \\
\checkmark & \texttimes & \texttimes & 0.78 & 1.5 & 1.8 \\
\texttimes & \texttimes & \texttimes & 0.75 & 1.8 & 1.8 \\
\bottomrule
\end{tabularx}
\end{subtable}
\hfill
\begin{subtable}[t]{0.48\linewidth}
\centering
\caption{PE/RoPE split and bin-count sensitivity.}
\label{tab:pe_rope_bins}

\newcommand{\triple}[3]{%
  \makebox[1.85em][r]{#1}%
  \hspace{0.25em}%
  \makebox[1.85em][r]{#2}%
  \hspace{0.25em}%
  \makebox[1.85em][r]{#3}%
}
\newcommand{\triplehead}{%
  \makebox[1.85em][c]{$R$}%
  \hspace{0.25em}%
  \makebox[1.85em][c]{$t$}%
  \hspace{0.25em}%
  \makebox[1.85em][c]{$d$}%
}

\begin{tabularx}{\linewidth}{@{}>{\raggedright\arraybackslash}Xc@{}}
\toprule
Configuration & ETH3D terrains \\
\midrule
Absolute PE only (no RoPE)
& \triple{0.68}{\textbf{0.9}}{\textbf{1.6}} \\
RoPE only (no Absolute PE)
& \triple{19.52}{7.8}{9.6} \\
Both (full)
& \triple{\textbf{0.48}}{\textbf{0.9}}{\textbf{1.6}} \\
\midrule
$N_r{=}10, N_\theta{=}8$
& \triple{0.72}{0.9}{1.7} \\
$N_r{=}20, N_\theta{=}8$ (default)
& \triple{0.48}{\textbf{0.9}}{1.6} \\
$N_r{=}40, N_\theta{=}8$
& \triple{\textbf{0.47}}{\textbf{0.9}}{\textbf{1.5}} \\
$N_r{=}20, N_\theta{=}0$ (radial only)
& \triple{2.82}{3.3}{3.6} \\
\bottomrule
\end{tabularx}
\end{subtable}

\end{table}

\section{Additional component ablations}
\label{sec:component_ablation_extended}

\cref{tab:extended_component_ablation} separates learned positional updates from parameter-free camera-model corrections.
On ETH3D terrains and FIORD Kitchen, learned PE/RoPE residuals are the dominant source of improvement.
Parameter-free corrections alone provide only small gains on ETH3D and degrade FIORD Kitchen, but they improve the learned adapter when combined with it.
This supports our claim that the learned positional residual is the main adaptation mechanism, while tokenization and prediction-grid corrections act as complementary camera model components.

\begin{table}[t]
\centering
\caption{\textbf{Learned positional residuals vs.\ parameter-free camera-model corrections.}
We report pose rotation error $R^\circ$, translation-direction error $t^\circ$, and reprojection error $d_{\mathrm{reproj}}$
on ETH3D terrains and FIORD Kitchen. Lower is better for all metrics.
\emph{Param.-free} denotes patch undistortion, prediction-grid coordinate correction, and border-token handling.
\textbf{Bold} marks the best value per dataset and metric.}
\label{tab:extended_component_ablation}
\small
\setlength{\tabcolsep}{4pt}
\renewcommand{\arraystretch}{1.05}

\begin{tabularx}{\linewidth}{@{}>{\raggedright\arraybackslash}p{0.15\linewidth}>{\raggedright\arraybackslash}Xccc ccc@{}}
\toprule
& & \multicolumn{3}{c}{ETH3D terrains}
& \multicolumn{3}{c}{FIORD Kitchen} \\
\cmidrule(lr){3-5}
\cmidrule(l){6-8}
Configuration & Details
& $R^\circ\downarrow$
& $t^\circ\downarrow$
& $d_{\mathrm{reproj}}\downarrow$
& $R^\circ\downarrow$
& $t^\circ\downarrow$
& $d_{\mathrm{reproj}}\downarrow$ \\
\midrule

Vanilla
& no PE/RoPE, no param.-free corrections
& 19.87 & 8.2 & 10.7
& 28.09 & 20.7 & 14.2 \\

Param.-free only
& patch undist., grid correct., border token
& 17.78 & 7.3 & 10.9
& 39.04 & 36.3 & 19.2 \\

Learned only
& PE+RoPE, no param.-free corrections
& 0.67 & 1.0 & 1.8
& 4.64 & 4.1 & 5.8 \\

\ourMethod Full
& PE+RoPE + param.-free corrections
& \textbf{0.48} & \textbf{0.9} & \textbf{1.6}
& \textbf{3.10} & \textbf{2.5} & \textbf{5.5} \\

\bottomrule
\end{tabularx}
\end{table}

\cref{tab:pe_rope_bins} separates absolute PE and RoPE adaptation on ETH3D terrains and evaluates the sensitivity to the number of radial ($N_r$) and angular ($N_\theta$) bins.
\emph{Absolute PE} is the main learned component on DA3-Small: using \emph{RoPE only} remains close to the unadapted model, while \emph{Absolute PE} alone nearly matches the full model.
The bin-count sensitivity shows that increasing the radial resolution beyond 20 bins gives only marginal gains, while removing angular bins substantially degrades performance. This indicates that the remaining camera-geometry mismatch contains an orientation-dependent component, not only a radial one.

\section{Broader impact}
\label{sec:broader_impact}
This work improves geometric perception for fisheye cameras, which are common in robotics, autonomous systems, AR/VR, and indoor mapping. More reliable geometry from wide-field cameras can lead to safer navigation and scene understanding, especially when standard pinhole assumptions fail. At the same time, like other 3D perception methods, improved reconstruction and pose estimation should be deployed with attention to privacy, consent, and safety in applications involving people or private spaces.
Because fisheye cameras are common in vehicles, mobile robots, doorbells, and indoor monitoring devices, easier camera adaptation can also lower the barrier to deploying 3D mapping systems in private or shared spaces. Responsible deployment should include consent, calibration validation, and application-specific safety checks.

\section{Existing assets and licenses}
\label{sec:licenses}
\cref{tab:asset_licenses} summarizes the external datasets, pretrained model weights, and code used in this work.
We use these assets only for non-commercial academic research and evaluation, and do not redistribute datasets or pretrained checkpoints.

\begin{table}[t]
\centering
\caption{\textbf{Existing assets used in this work.} External datasets and pretrained models used for evaluation or adaptation.}
\label{tab:asset_licenses}
\small
\setlength{\tabcolsep}{3pt}
\begin{tabularx}{\linewidth}{ll l X}
\toprule
Asset & Type & Use & License / terms \\
\midrule
KITTI-360~\citep{liao2022kitti360noveldatasetbenchmarks}
& Dataset & Evaluation
& CC BY-NC-SA 3.0; scripts MIT \\

TUM-VI~\citep{Schubert_2018}
& Dataset & Evaluation
& Data CC BY 4.0; code BSD-2-Clause \\

ScanNet++~\citep{yeshwanth2023scannethighfidelitydataset3d}
& Dataset & Evaluation
& Terms of Use; non-commercial research/education \\

ETH3D~\citep{schops2017multi}
& Dataset & Evaluation
& CC BY-NC-SA 4.0 \\

FIORD~\citep{gunes2025fiordfisheyeindooroutdoordataset}
& Dataset & Evaluation
& CC BY 4.0 \\

DA3~\citep{lin2025depth}
& Model/code & Backbone
& Apache-2.0; CC BY-NC 4.0 for Large/Giant \\

Pi3~\citep{wang2026pi3permutationequivariantvisualgeometry}
& Model/code & Backbone
& Code BSD-3-Clause; weights CC BY-NC 4.0 \\

VGGT~\citep{wang2025vggt}
& Model/code & Backbone
& Checkpoint-specific license; non-commercial checkpoint \\

UFM~\citep{zhang2025ufm}
& Model/code & Matches
& Code BSD-3-Clause; checkpoint CC BY-NC-SA 4.0 \\
\bottomrule
\end{tabularx}
\end{table}

\end{document}